\documentclass[acmsmall,screen]{acmart}

\usepackage{amsmath,amsfonts,amssymb}
\usepackage{algorithmic}
\usepackage{graphicx}
\usepackage{textcomp}
\usepackage{multirow}
\usepackage{multicol}
\usepackage{xcolor}
\usepackage{hyperref}       
\usepackage{url}            
\usepackage{booktabs}       
\usepackage{nicefrac}       
\usepackage{microtype}      
\usepackage{bbding}
\usepackage{booktabs}
\usepackage{bbm}
\usepackage{colortbl}
\usepackage{changepage} 
\usepackage{caption}
\usepackage{placeins}
\usepackage{wrapfig}
\usepackage{floatflt}
\usepackage{enumitem}
\usepackage{stfloats}
\usepackage{lipsum}
\usepackage{makecell}
\usepackage{bm}
\usepackage{subcaption} 

\newcounter{rownumber}
\newcommand{\rownumber}{\stepcounter{rownumber}\arabic{rownumber}}

\AtBeginDocument{%
  }

\setcopyright{acmlicensed}
\copyrightyear{2025}
\acmYear{2025}
\acmDOI{XXXXXXX.XXXXXXX}
\acmISBN{978-1-4503-XXXX-X/2018/06}

\begin{document}

\title{Re-purposing SAM into Efficient Visual Projectors for MLLM-Based Referring Image Segmentation}

\author{Xiaobo Yang}
\email{hal_42@zju.edu.cn}
\orcid{0009-0003-7885-302X}
\author{Xiaojin Gong}
\authornote{Corresponding Author}
\email{gongxj@zju.edu.cn}
\orcid{0000-0001-9955-3569}
\affiliation{%
  \institution{Zhejiang University}
  \city{Hangzhou}
  \state{Zhejiang}
  \country{China}
}

\renewcommand{\shortauthors}{Yang et al.}

\begin{abstract}
Recently, Referring Image Segmentation (RIS) frameworks that pair the Multimodal Large Language Model (MLLM) with the Segment Anything Model (SAM) have achieved impressive results.
However, adapting MLLM to segmentation is computationally intensive, primarily due to visual token redundancy.
We observe that traditional patch-wise visual projectors struggle to strike a balance between reducing the number of visual tokens and preserving semantic clarity, often retaining overly long token sequences to avoid performance drops.
Inspired by text tokenizers, we propose a novel semantic visual projector that leverages semantic superpixels generated by SAM to identify ``visual words" in an image.
By compressing and projecting semantic superpixels as visual tokens, our approach adaptively shortens the token sequence according to scene complexity while minimizing semantic loss in compression.
To mitigate loss of information, we propose a semantic superpixel positional embedding to strengthen MLLM's awareness of superpixel geometry and position, alongside a semantic superpixel aggregator to preserve both fine-grained details inside superpixels and global context outside.
Experiments show that our method cuts visual tokens by $\sim$93\% without compromising performance, notably speeding up MLLM training and inference, and outperforming existing compressive visual projectors on RIS.
\end{abstract}



\received{20 February 2007}
\received[revised]{12 March 2009}
\received[accepted]{5 June 2009}

\maketitle

\section{Introduction}
\label{sec:Introduction}
Referring Image Segmentation (RIS)~\cite{10.1145/3698771,10.1145/3701733,XIE2024111243,li2021referring,lai2023lisa,hanoona2023GLaMM} requires segmenting the object within an image that matches a given textual description. In contrast to traditional image segmentation, which is limited to a predefined label set, RIS lets users locate objects through free-form natural language, using attributes, spatial relations, and other fine-grained cues to distinguish among objects of the same class. This capability is highly valuable for tasks such as image editing and robotic perception.
RIS remains one of the most challenging problems in segmentation and multimodal understanding because it demands nuanced comprehension of text, vision, and their cross-modal interactions.

Multimodal Large Language Models (MLLMs), pre-trained on large-scale multimodal data, exhibit powerful vision-language understanding. 
To transfer this capability for pixel-level grounding, recent studies~\cite{lai2023lisa,hanoona2023GLaMM,xu2024ullava,yuan2025sa2va} fine-tune MLLMs on RIS datasets so that, given a referring expression, the model can prompt SAM~\cite{Kirillov2023SAM} with the hidden state of a mask-related token to produce the segmentation mask.

\begin{figure}[t]
     \centering
     \begin{subfigure}
          [t]{1.0\linewidth}
          \centering
          \includegraphics[width=\linewidth]{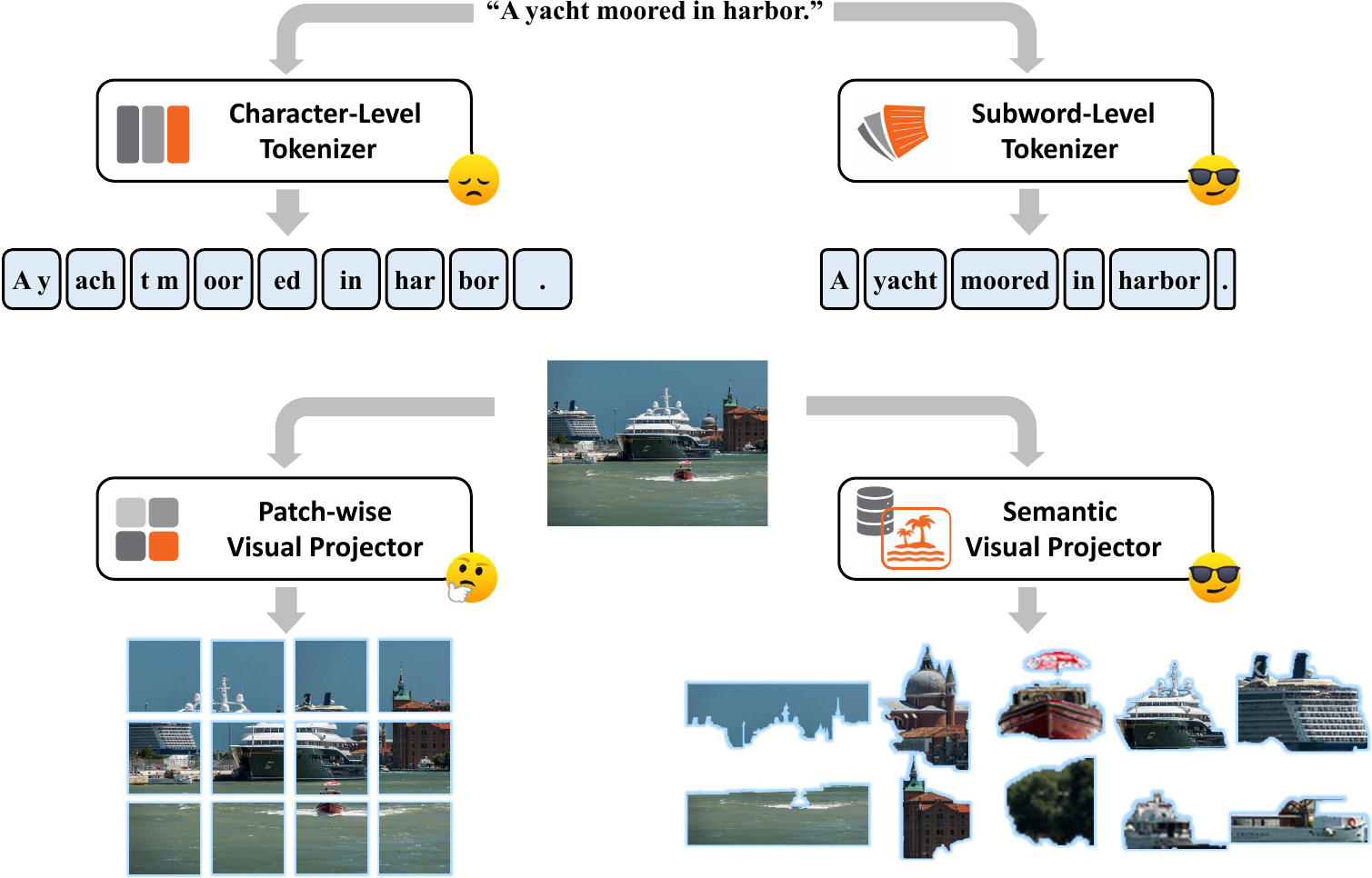}
     \end{subfigure}
     \vskip 0.5em
     \begin{subfigure}
          [t]{1.0\linewidth}
          \centering
          \includegraphics[width=\linewidth]{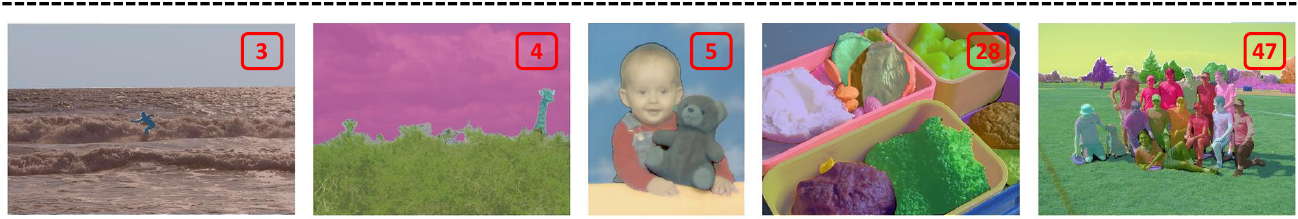}
     \end{subfigure}
     \caption{ Our motivation.
     Top: Unlike LLMs that tokenize text based on subwords, images are often split into fixed-size patches, disregarding visual semantics—akin to grouping text by a fixed number of characters.
     This trade-off between large, semantically blurry patches and small, computationally expensive patches motivates our approach: encoding semantic superpixels as visual tokens to adaptively reduce token number while preserving semantic integrity.
     Note that this figure highlights the limitations of large patches used by compressive visual projectors, whereas MLP projectors use smaller but denser patches (e.g., 576).
     Bottom: Examples of semantic superpixels. The red number
     marks the superpixel count, which scales adaptively with scene complexity. }
     \Description{}
     \label{fig:Sec5/trigger}
\end{figure}

Although MLLM-based RIS frameworks achieve impressive performance, their large number of visual tokens presents a significant efficiency bottleneck that slows down training and inference.
Typically, a MLLM comprises:
(1) a vision encoder that divides an image into equally sized square patches and encodes them as patch-wise image embeddings;
(2) a patch-wise visual projector that maps patch-wise embeddings into the LLM embedding space as visual tokens; and
(3) a Large Language Model (LLM) that processes the combined visual and text tokens for understanding and generation.
The parameter-heavy LLM dominates the overall runtime cost~\cite{chen2024image}, and its computational complexity grows quadratically with the input token length.
In MLLM-based RIS frameworks, that token budget is primarily consumed by visual tokens. For example, a single dialogue of GLaMM~\cite{hanoona2023GLaMM} includes about 60 text tokens but up to 576 visual tokens.

We begin by re-examining widely adopted patch-wise visual projectors and pointing out that they face a trade-off between two competing issues:
(1) Semantic redundancy. Due to the low information density of natural images, many neighboring patches convey the same meaning, forcing the LLM to waste computation on redundant information.
For the image in Fig.~\ref{fig:Sec5/trigger}, dozens or even hundreds of adjacent tokens may all correspond to the sky or water.
(2) Semantic ambiguity.
While increasing the patch size can cut the token number and reduce redundancy, it risks blending regions with different semantics, causing loss of semantics or confusion within visual tokens (e.g., the ``boats" and ``buildings" in Fig.~\ref{fig:Sec5/trigger}).

Existing patch-wise visual projectors inevitably fall on one side of the trade-off.
The most commonly used MLP maps high-resolution patch-wise embeddings to visual tokens one-to-one, producing a semantically clear yet redundant token sequence.
To reduce token number, some compressive visual projectors~\cite{chu2023mobilevlm,cha2023honeybee,li2024tokenpacker,yao2024deco,lu2025internvl} downsample patch-wise image embeddings, i.e., merging neighboring patches into larger ones, resulting in semantically ambiguous tokens and markedly degrading RIS performance.
Moreover, these projectors rely on a preset downsampling rate and, therefore, cannot adapt the sampling density to match the semantic complexity of each image.

This paper proposes a Semantic Visual Projector (SVP), which maps semantic superpixels to a compressed set of visual tokens, dramatically reducing the token number while preserving semantic clarity.
Our design draws inspiration from the tokenizers of Natural Language Processing (NLP).
Segmenting text by fixed-length characters inflates token number and breaks morphological or syntactic boundaries, so modern NLP adopts statistical algorithms such as BPE~\cite{sennrich2015neural}, which split text into semantic elements (referred to as ``subwords") that balance brevity and semantic integrity.
An analogous mechanism exists in vision. FMA-WSSS~\cite{FMA-WSSS} leverages SAM~\cite{Kirillov2023SAM} to partition images into regions of internal semantic consistency called semantic superpixels.
With SAM exploited statistical regularities in large-scale image data, semantic superpixels can capture common objects by a single high-level SAM mask, whereas rarer or more complex regions are split into several lower-level masks.
SVP thus treats each superpixel as a “visual word” and maps it to a single visual token.
This semantics-aware compression offers a concise, comprehensive, and clear representation of the image, reconciling speed and accuracy.
Besides, SVP naturally adapts to scene complexity: it assigns more tokens to complex scenes to capture fine detail and fewer to simpler scenes to maximize efficiency.

To equip the MLLM with image structural cues, we further propose the Semantic Superpixel Positional Embedding (SSPE), encoded via an SSPE encoder.
Since semantic superpixels vary in shape and size and lack an intrinsic order to represent their relative positions, the default Rotary Position Embedding (RoPE)~\cite{su2024roformer} fails to convey their geometry or spatial relationships.
We address this issue by introducing the SSPE encoder that uses multiple learnable queries to adaptively encode superpixels' geometric and positional information into SSPE, which are then injected into visual tokens.

Finally, we introduce the Semantic Superpixel Aggregator (SSA) to better distinguish the target object from others of the same class. 
By sharing SSPE at the superpixel level in cross-attention layers, the SSA simultaneously enhances local semantics and integrates surrounding environmental context. 
This mechanism refines the object's fine-grained attributes and captures its interaction with the environment, providing discriminative cues for robust identification.

Our contributions can be summarized as follows:
\begin{itemize}
    \item We identify a critical trade-off in widely used patch-wise visual projectors between semantic redundancy and ambiguity.
    To address this, we propose a Semantic Visual Projector (SVP), which compresses visual inputs by aggregating semantic superpixels into semantically coherent tokens.
    SVP adapts the token number to scene complexity, cutting it by $\sim$93\% on average while preserving semantic integrity.

    \item We propose a Semantic Superpixel Positional Embedding (SSPE) encoder that adaptively encodes superpixel geometry and position using multiple learnable queries, generating SSPE to enhance the MLLM's understanding of the image structure.

    \item We propose a Semantic Superpixel Aggregator (SSA) to jointly model semantics inside superpixels and outside context by cross-attention layers with SSPE shared at the superpixel level. 

    \item Experiments on the RIS show that SVP significantly improves the training and inference efficiency of MLLMs compared to the standard MLP projector while maintaining comparable performance.
    Under the same compression ratio, SVP substantially outperforms existing compressive visual projectors.
\end{itemize}

\section{Related Work}
\label{sec:related_work}

\subsection{Multimodal Large Language Models}
MLLMs employ a visual projector to bridge an advanced vision encoder with an LLM, endowing the system with strong cross-modal understanding and generation capabilities.
Early efforts like Flamingo~\cite{alayrac2022flamingo} and BLIP-2~\cite{li2023blip2} showed that paired image–text data can successfully adapt LLMs to visual input.
Subsequent studies—including MiniGPT-4~\cite{chen2023minigptv2,zhu2023minigpt}, Instruct-BLIP~\cite{instructblip}, the LLaVA~\cite{liu2023improvedllava,liu2023llava}, InternVL~\cite{gao2024mini,chen2024far,chen2024internvl}, and Qwen-VL~\cite{Qwen-VL,Qwen2-VL,Qwen2.5-VL} families—continuously advanced MLLM capabilities by adopting more powerful vision encoders and LLM, enlarging datasets, refining training pipelines, and enabling dynamic-resolution.
To move beyond the limits of text-only output, some methods~\cite{wang2023visionllm,wu2024visionllm,lai2023lisa,xu2024ullava} pair MLLMs with external tools such as Stable Diffusion~\cite{podell2023sdxl} or SAM~\cite{Kirillov2023SAM}, unlocking downstream tasks like image editing and visual grounding.
Nevertheless, the visual projector of MLLMs remains under-explored.
In this paper, we focus on developing a projector that strikes a balance between efficiency and accuracy for pixel-level tasks.

Several studies further aim to enable MLLMs to interact with the Region of Interest (RoI).
For RoIs defined by bounding boxes, GPT4RoI~\cite{zhang2023gpt4roi} encodes semantics of RoI with RoI-aligned features, whereas others~\cite{xuan2024pink,chen2023shikra,peng2023kosmos,youferret} encode RoI coordinates as text or discrete location tokens.
Ferret~\cite{youferret} further introduces a spatial-aware visual sampler to obtain visual features for arbitrarily shaped regions.
However, these methods all input patch-wise visual tokens first and then reference regions, so the visual token number stays high.
Furthermore, referring to a specific region with text or location tokens usually costs several additional tokens.
We instead represent an image as a set of region-specific visual tokens, assigning exactly one token per region and thus drastically reducing the token number.

\subsection{Referring Image Segmentation}
Early RIS approaches~\cite{li2021referring,wang2022cris,yang2022lavt,liu2023polyformer} typically employ text encoders~\cite{devlin-etal-2019-bert} to embed referring expressions, which then guide the segmentation.
X-Decoder~\cite{zou2022xdecoder} and SEEM~\cite{zou2023segment} extend this idea by employing a unified decoder that treats expressions as textual prompts, thereby casting RIS as a promptable segmentation task.
Vision foundation models have also been widely applied to RIS. For example, Grounded-SAM~\cite{ren2024grounded} first uses the open-vocabulary detector Grounding-DINO~\cite{GroundingDINO} to predict bounding boxes, which then serve as prompts for SAM~\cite{Kirillov2023SAM} to obtain the final mask.

More recently, MLLMs have been leveraged to jointly encode images and referring expressions, capitalizing on their strong cross-modal reasoning capabilities acquired through large-scale pre-training.
PerceptionGPT~\cite{pi2024perceptiongpt}, PixelLM~\cite{ren2024pixellm}, and OMG-LLaVA~\cite{OMGLLaVA} train a decoder on top of MLLM outputs to generate segmentation masks.
F-LMM~\cite{wu2024flmm} instead exploits the geometric and spatial cues already present in the MLLM’s self-attention to predict a mask, which is further refined by a decoder-tunable SAM.
LISA~\cite{lai2023lisa} and GLaMM~\cite{hanoona2023GLaMM} fine-tune LLaVA~\cite{liu2023improvedllava,liu2023llava} so that, given a natural-language instruction, the model can prompt a decoder-learnable SAM to produce the segmentation result.
Building on LISA, M$^{2}$SA~\cite{jang2025mmr} introduces early local feature fusion and multiple [SEG] tokens, enabling more effective multi-target and multi-granularity reasoning segmentation.
While incorporating MLLMs has notably improved RIS performance, it has also greatly increased training and inference costs. Therefore, this paper focuses on boosting the efficiency of adapting MLLMs to RIS.

\subsection{Visual Token Reduction}
Widely used MLP visual projectors map patch-wise image embeddings to visual tokens in a one-to-one manner, retaining all image details.
Nevertheless, the resulting redundant visual tokens have become a major efficiency bottleneck for MLLMs, encouraging researchers to reduce the computational load over visual tokens.

Various importance-based methods have emerged, including locating unimportant visual tokens within the vision encoder or LLM for pruning, merging, or skipping computations~\cite{LIANG2024111664,li2025qg,chen2024image,zhang2024sparsevlm,wen2025stop,li2025beyond,zeng2025skip}, identifying the most significant image embeddings from vision encoder outputs~\cite{zhang2024cls,shang2024llava}, and lowering the resolution of visual tokens in certain LLM layers~\cite{lu2025internvl}.
However, these methods rely on hand-crafted heuristics to evaluate token importance, which may not generalize well across different tasks, datasets, or models~\cite{wen2025stop,zhang2024cls,lu2025internvl}.
Moreover, some methods require access to or modification of the internal mechanisms of transformers, raising compatibility issues with acceleration techniques like KV-Cache or FlashAttention~\cite{dao2022flashattention,dao2023flashattention2}.

An alternative research line designs compressive visual projectors that adaptively compress patch-wise image embeddings into visual tokens without relying on hand-crafted heuristics to evaluate token importance or modifying the computation pipelines of the vision encoder or LLM.
Early methods such as Resampler~\cite{Qwen-VL} and Q-Former~\cite{li2023blip2,instructblip} use global cross-attention to aggregate image embeddings into a fixed number of queries. However, due to a lack of inductive bias regarding location, these projectors' queries tend to revisit the most salient areas of the image repeatedly, neglecting local details~\cite{cha2023honeybee,lu2025internvl}.
Therefore, more recent compressive visual projectors primarily utilize downsampling to create compressed visual tokens by partitioning image patches into rectangular windows and merging local embeddings within them.
These methods can also be classified as patch-wise visual projectors because the windows can be viewed as larger patches.
Some studies explore classical downsampling modules such as pixel-shuffle~\cite{chen2024internvl} and adaptive average pooling~\cite{yao2024deco}, while others design more complex modules to mitigate detail loss.
For example, Honeybee~\cite{cha2023honeybee} employs convolution or deformable attention with average pooling for downsampling image embeddings, while LDPv2~\cite{chu2023mobilevlm,chu2024mobilevlm} stacks MLPs, average pooling, and convolution layers for this purpose.
TokenPacker~\cite{li2024tokenpacker} and PVTC~\cite{lu2025internvl} perform cross-attention within the local window to refine the coarse embeddings downsampled by bilinear interpolation or pixel-shuffle.
Nonetheless, downsampling-based projectors inevitably produce semantically ambiguous tokens due to enlarged patch sizes, and their downsampling rates are preset rather than dynamically adjusted based on scene content. In contrast, our SVP maps semantic superpixels rather than image patches to visual tokens, thereby avoiding these issues.

\section{Preliminary}

\subsection{Semantic Superpixel}
Prompting SAM~\cite{Kirillov2023SAM} with a grid of points yields a set of high-quality SAM masks that aim to capture visual elements throughout the image and align closely with their boundaries.
However, each prompt point will produce three nested masks, corresponding to the \textit{whole}, \textit{part}, and \textit{sub-part} of objects, which results in substantial overlap among SAM masks and yields an average of up to 150 masks per image.

To minimize both mask overlap and mask quantity, we adopt the filtering algorithm from FMA-WSSS~\cite{FMA-WSSS} to select a subset of SAM masks referred to as semantic superpixels (also known as SAM-based quasi-superpixels).
This method prioritizes larger \textit{whole} masks while discarding overlapping subordinate masks, thereby minimizing the total number of masks required to cover the entire image.
When no \textit{whole} mask is available for a region, \textit{part} masks are selected, and finally \textit{sub-part} masks are used to ensure that all visual elements across the image are captured.

As a result, the image is decomposed into an average of 40 fundamental visual elements by semantic superpixels, which provide a complete and non-redundant representation.
The semantics within each semantic superpixel are consistent and coherent, allowing for a clear and effective representation through visual tokens.
In contrast, further subdivision  (e.g., into smaller patches) would introduce unnecessary redundancy.
Additionally, similar to the behavior of BPE~\cite{sennrich2015neural}, which prioritizes high-frequency words as single subwords, semantic superpixels often capture common objects in one shot by ``whole" masks, while subordinate masks are used to decompose rarer scenes. Therefore, SVP can adaptively adjust the number of visual tokens based on scene complexity.
We visualize and compare semantic superpixels with traditional superpixels in Sec.~\ref{sec:slic_vs_sam}.

\subsection{GLaMM}
We evaluate our SVP primarily under the framework of GLaMM~\cite{hanoona2023GLaMM}, which is a state-of-the-art MLLM-based RIS framework.
GLaMM fine-tunes LLaVA~\cite{liu2023improvedllava,liu2023llava} to perform Grounded Conversation Generation (GCG), enabling it to segment objects within input images according to natural language instructions.
For example, when a user provides an image and asks, ``Can you segment the yacht moored in harbor?" GLaMM responds with ``It is [SEG]." Here, the hidden embedding corresponding to the special token [SEG] is decoded into the segmentation mask of the ``yacht".

GLaMM follows LLaVA by using a frozen CLIP-L~\cite{Radford2021CLIP} as the vision encoder and an MLP as the visual projector. The hidden embedding at the [SEG] token is projected into a prompt vector through another MLP (referred to as the L-P projector).
Meanwhile, the input image is encoded by the frozen SAM~\cite{Kirillov2023SAM} encoder, whose output, together with the prompt vector, is fed into a trainable SAM decoder to generate the segmentation mask.
During fine-tuning on the RIS task, GLaMM is supervised by autoregressive loss on the output text and a set of segmentation losses on the generated masks. The LLM is fine-tuned with LoRA~\cite{hu2022lora}, while the vision encoder and SAM encoder remain frozen.

\section{The Proposed Method}


\begin{figure}[t]
    \centering
    \includegraphics[width=1.0\linewidth]{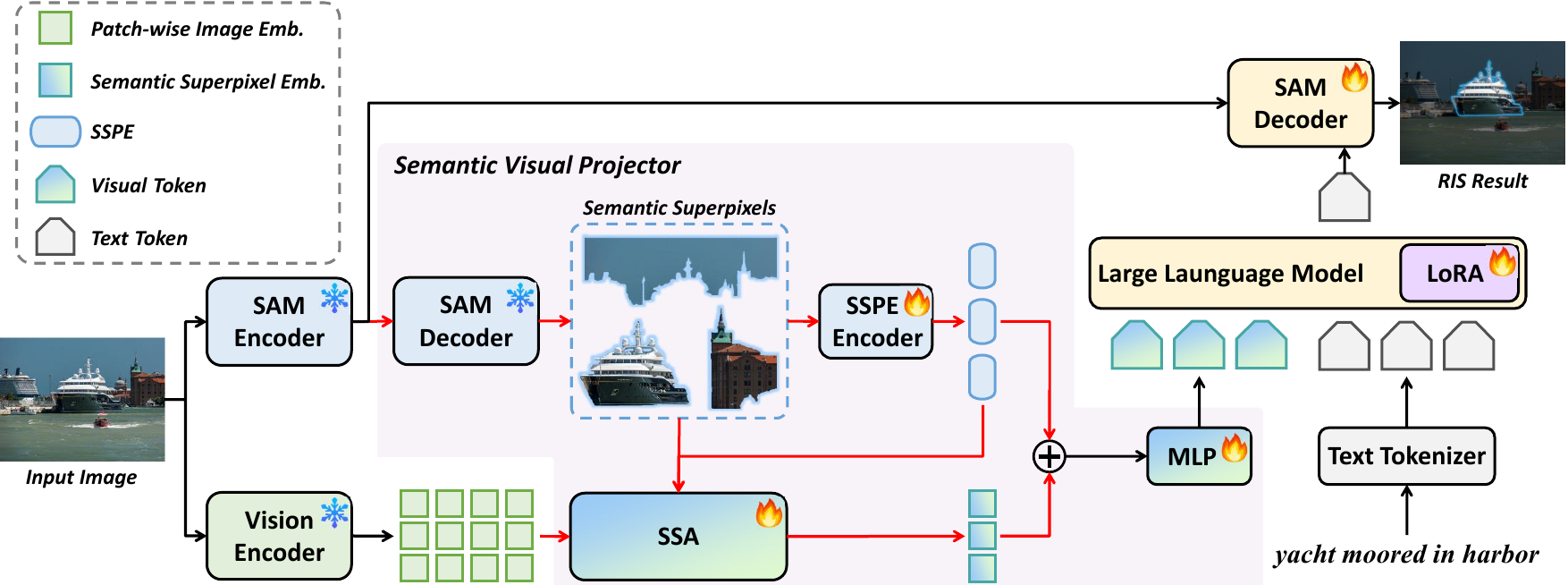}
    \caption{ Method overview.
        GLaMM~\cite{hanoona2023GLaMM} Pipeline (Black Arrows):
        The vision encoder encodes the image into patch-wise embeddings, which an MLP then projects into visual tokens in a one-to-one manner.
        These tokens, along with text tokens, are fed into the LLM, whose response is decoded by a decoder-learnable SAM into the RIS result.
        SVP Pipeline (Red Arrows):
        Semantic superpixels, obtained via a frozen SAM decoder, are encoded into SSPE.
        The SSA aggregates image embeddings into semantic superpixel embeddings, which are added with SSPE and projected into compressed visual tokens.
    }
    \Description{}
    \label{fig:Sec5/overview}
\end{figure}

\subsection{Overview}

As illustrated in Fig.~\ref{fig:Sec5/overview}, we apply our Semantic Visual Projector (SVP) to RIS with the following modifications to the network architecture of GLaMM~\cite{hanoona2023GLaMM}:

(1) The output of the SAM encoder is reused and fed into a frozen SAM decoder to generate semantic superpixels.
This step adds only a single inference pass through a lightweight SAM decoder, resulting in negligible overhead.

(2) The patch-wise MLP visual projector was replaced by SVP.
Specifically, the patch-wise image embeddings of resolution $H \times W$ are reshaped into $\mathbf{F} \in \mathbb{R}^{HW \times d}$.
$M$ semantic superpixels $\mathbf{S} \in \mathbb{B}^{M \times HW}$ are encoded into Semantic Superpixel Positional Embeddings (SSPE) $\mathbf{P}_{sp} \in \mathbb{R}^{M \times d}$ by an SSPE encoder.
The Semantic Superpixel Aggregator (SSA) then aggregates image embeddings according to semantic superpixels, producing semantic superpixel embeddings $\mathbf{E} \in \mathbb{R}^{M \times d}$. $\mathbf{E}$ are then added to the SSPE and projected into visual tokens $\mathbf{T}_v \in \mathbb{R}^{M \times d'}$ via an MLP, formulated as:
\begin{equation}
    \mathbf{T}_v = \operatorname{MLP}(\mathbf{E} + \mathbf{P}_{sp}),
\end{equation}
where $d'$ is the input dimension of the LLM.

(3) We find that the CLIP-L vision encoder, limited by image-level pre-training, tends to overlook the attribute information in non-salient regions, resulting in the compressed visual tokens lacking fine-grained semantics required for RIS.
To address this issue, we replace the vision encoder with DINOv2~\cite{dinov2}, a self-supervised representation learning model that tends to retain attribute information across arbitrary locations.
The necessity of this replacement is discussed in detail in Sec.~\ref{sec:replace_visual_encoder}.

\begin{figure}[t]
    \centering
    \includegraphics[width=1.0\linewidth]{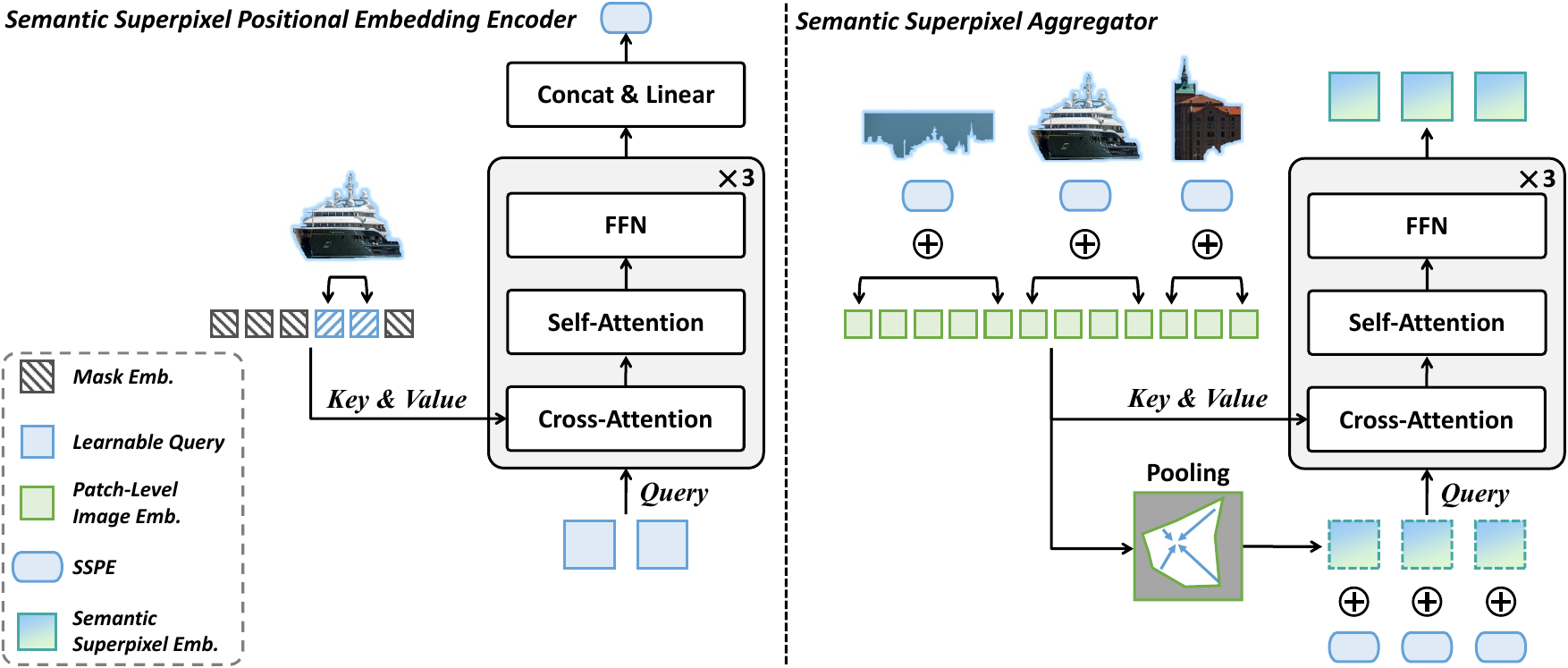}
    \caption{
        Left: Illustration of the SSPE encoder.
        Right: Illustration of the SSA.
        For clarity, the 2D sinusoidal positional embedding~\cite{vaswani2017attention} added to the mask and image embeddings is omitted.
        Layer normalization and residual connections for the cross-attention, self-attention, and feed-forward layers (FFN) are also omitted.
        Positional embeddings are re-added to the query and key at each cross-attention or self-attention layer.
    }
    \Description{}
    \label{fig:Sec5/SSA}
\end{figure}

\subsection{Semantic Superpixel Positional Embedding}

The LLM captures the order of tokens by adding sequential positional embeddings (such as RoPE~\cite{su2024roformer}) to token sequences.
For visual tokens derived from uniform patches, feeding them into the LLM in raster order effectively conveys their relative positions and the spatial structure of the image.

However, semantic superpixels can vary significantly in shape and size.
While traditional positional embeddings assume patches are equally sized squares, they fail to account for the geometric properties of the regions corresponding to visual tokens.
Moreover, the spatial relationships between semantic superpixels cannot be accurately represented by the token order.
For instance, in Fig.~\ref{fig:Sec5/trigger}, the semantic superpixel covering a ``person" is surrounded by the ``sea" superpixel, making it difficult to establish a clear precedence between them.
Therefore, sequential positional embeddings are inadequate for representing the image structure formed by semantic superpixels, hindering RIS related to shape, size, or location.

To enhance MLLM's spatial awareness of irregular regions, we design a Semantic Superpixel Positional Embedding (SSPE) encoder that adaptively captures superpixels' geometry and position using multiple learnable queries to generate enriched positional embeddings.
As shown in Fig.~\ref{fig:Sec5/SSA}, the SSPE encoder comprises three identical blocks, each sequentially incorporating cross-attention, self-attention, and feed-forward layers.
The semantic superpixels represented by binary masks are first encoded into mask embeddings by an embedding layer, which serve as keys and values for all cross-attention layers.
Meanwhile, $N$ learnable queries are input into the encoder to capture the geometry and position of the masks.
The outputs of these queries are concatenated and linearly projected to obtain the SSPE for each superpixel. Stacking the SSPEs for $M$ superpixels in the image produces $\mathbf{P}_{sp} \in \mathbb{R}^{M \times d}$, which is integrated into visual tokens through addition.
Before entering the attention layer, the mask embeddings are added to 2D sinusoidal positional embeddings~\cite{vaswani2017attention}, while the queries are added to learnable positional embeddings of the same shape.

Our SSPE encoder's design is inspired by the classical encoder-decoder architecture~\cite{vaswani2017attention,raffel2020exploring}, which has proven effective for spatial feature extraction by object detection models like DETR~\cite{carion2020end}. 
Whereas DETR predicts bounding box coordinates from complex image embeddings, our encoder is tailored to extract superpixel shape, size, and position from simpler binary mask inputs.
Unlike DETR, the SSPE encoder assigns multiple queries to each superpixel, facilitating a more comprehensive capture of geometric information.
Furthermore, unlike methods that encode region coordinates as additional tokens~\cite{xuan2024pink,chen2023shikra,peng2023kosmos,youferret}, SSPE is instead added to the token sequence.
This design yields a detailed, data-driven representation of superpixel geometry and position without introducing extraneous tokens.

\subsection{Semantic Superpixel Aggregator}
We propose the Semantic Superpixel Aggregator (SSA) to aggregate patch-wise image embeddings into semantic superpixel embeddings, which effectively integrates the internal semantics of superpixels with the relevant context.
As shown in Fig.~\ref{fig:Sec5/SSA}, the structure of the SSA is similar to the SSPE encoder.
The keys and values of cross-attention layers derive from the image embedding $\mathbf{F} \in \mathbb{R}^{HW \times d}$.
Simultaneously, the semantic superpixels $\mathbf{S} \in \mathbb{B}^{M \times HW}$ apply average pooling to the covered image embeddings, producing coarse semantic superpixel embeddings $\mathbf{E}' \in \mathbb{R}^{M \times d}$:
\begin{equation}
    \label{eq:avg_pool_e}
    \mathbf{E}' = \mathcal{N}_1(\mathbf{S}) \mathbf{F},
\end{equation}
where $\mathcal{N}_1(\cdot)$ denotes $L_1$ normalization along the last dimension.
The $\mathbf{E}'$ serves as the query input for the SSA.
In the SSA, the coarse embeddings, obtained via average pooling, resample the most discriminative image embeddings within the superpixels, fully exploring the fine-grained attributes of the object while integrating contextual knowledge, ultimately yielding the semantic superpixel embedding $\mathbf{E}$.

Referring again to Fig.~\ref{fig:Sec5/SSA}, the query's positional embedding is the Semantic Superpixel Positional Embedding (SSPE) of its corresponding semantic superpixel.
When used as keys, $\mathbf{F}$ is added with two sets of positional embeddings: (1) a two-dimensional sinusoidal positional embedding~\cite{vaswani2017attention} to differentiate between patches, and (2) $\mathbf{P}_{patch} \in \mathbb{R}^{HW \times d}$ to distinguish between semantic superpixels, computed as:
\begin{equation}
    \label{eq:mask_pos}
    \mathbf{P}_{patch} = \mathbf{S}^\top \mathbf{P}_{sp},
\end{equation}
meaning that the SSPE for each semantic superpixel is added to the patch-wise image embeddings it covers.

\begin{figure}[t]
    \centering
    \includegraphics[width=0.8\linewidth]{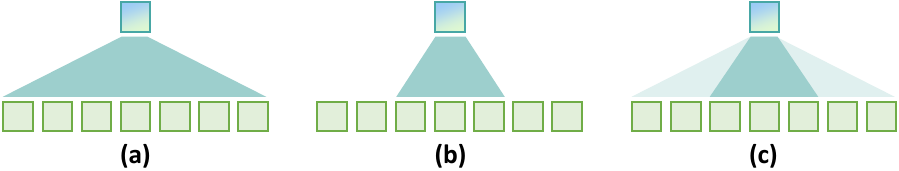}
    \caption{
        Illustration of different cross-attention mechanisms.
        (a) The query attends to all image embeddings but lacks local priors, like Q-Former~\cite{li2023blip2,instructblip}.
        (b) The query can only attend to a specific region, like Mask2Former~\cite{mask2former}.
        (c) The query prioritizes focuses on the semantic superpixel it belongs to, while also accessing a broader context, like our SSA.
    }
    \Description{}
    \label{fig:Sec5/sspe_share}
\end{figure}

The SSA adapts the concept of sharing positional embeddings for queries and keys of the same patch from ViT~\cite{dosovitskiy2020image} to the semantic superpixel level, where the query and image embeddings for the same superpixel share the SSPE.
This superpixel-level sharing encourages superpixel embeddings to (1) prioritize capturing and refining intra-superpixel details and (2) adaptively incorporate fine-grained contextual information from outside.
This design directly contrasts with other cross-attention mechanisms.
As shown in Fig.~\ref{fig:Sec5/sspe_share}, queries of Q-Former~\cite{li2023blip2,instructblip} lack local priors, making them prone to overlook less-salient regions or details.
Additionally, instead of rigidly constraining the attention area to specific regions through attention biases or reshaping~\cite{mask2former,li2024tokenpacker,lu2025internvl}, the SSA allows for broader contextual access, enabling MLLMs to better resolve relational or comparative references (e.g., ``the fork on the plate," ``the taller giraffe").

\section{Experiments}

\subsection{Datasets and Evaluation Metrics}

We train our model following GLaMM~\cite{hanoona2023GLaMM} using a mixture of training sets from three mainstream RIS datasets: RefCOCO, RefCOCO+~\cite{refcoco}, and RefCOCOg~\cite{refcocog}, evaluating performance on their respective validation or test sets.
RefCOCO and RefCOCO+ (using the ``unc" split) have a validation set containing 1,500 images and two test sets, \textit{testA} and \textit{testB}, each containing 750 images focused on people and non-people, respectively.
RefCOCOg (using the ``umd" split) contains 1,300 validation and 2,600 test images, featuring richer object descriptions that challenge language comprehension.
Additionally, RefCOCO+ disallows location words in the expressions (e.g., ``man in the yellow polka-dotted shirt" rather than ``man to the right"), forcing the model to distinguish same-class objects based on their appearance and context, thereby requiring advanced fine-grained visual understanding.

Common RIS metrics include cumulative IoU (cIoU) and generalized IoU (gIoU)~\cite{liu2023gres}.
cIoU computes the Intersection over Union (IoU) for the foreground across all pixels in the dataset but is easily biased by large objects.
The gIoU avoids this by averaging the IoU for each image, ensuring all samples are weighted equally.
Besides, cIoU is heavily penalized by False Positives (FP) with larger unions, whereas gIoU is equally sensitive to both False Negatives (FN) and FP.
Since visual tokens from compressive visual projectors cover larger areas than those from the MLP, they increase the likelihood of FP and \textbf{unfairly} disadvantage our model under the cIoU.
Thus, we adopt gIoU as the default metric unless otherwise stated.

\subsection{Implementation Details}

We follow FMA-WSSS~\cite{FMA-WSSS} to generate semantic superpixels and randomly shuffle their order during training, forcing the MLLM to rely on SSPE instead of unreliable orderings to understand the image structure.
The semantic superpixels input to the SSPE encoder are adaptively interpolated by a factor of $\sqrt{40 \times 126^2 / (MHW)}$, yielding a total pixel number of approximately $40 \times 126^2$, which balances high mask resolution with stable memory usage.
We use SAM-Fast~\cite{SAM-Fast} to accelerate SAM's computation.
The DINOv2~\cite{dinov2} vision encoder employs a ViT-L/14 backbone with registers~\cite{dosovitskiy2020image}, taking the output from its last layer as patch-wise image embeddings.
The implementation of the cross-attention, self-attention, and feed-forward layers for the SSPE encoder and the SSA is consistent with Mask2Former~\cite{mask2former}.
Given that the input to the SSPE encoder is simple binary masks, the embedding dimension of its multi-head attention is set to only 128, with 8 heads to improve efficiency.
We use $N=4$ learnable queries for SSPE.
The embedding dimension and number of heads for the SSA are kept consistent with the vision encoder.

During training, we load the parameters of GLaMM pre-trained on the GranD dataset~\cite{hanoona2023GLaMM}, while the SVP is randomly initialized.
The vision encoder, SAM encoder, and SAM decoder for generating semantic superpixels remain frozen, and we fine-tune the SVP, LLM's embedding layer and projection head, L-P projector, and SAM decoder for generating RIS masks.
The LLM is fine-tuned by LoRA~\cite{hu2022lora} with a rank of 32 and $\alpha$ of 16.
Training is conducted in two stages: First, the LLM is frozen and trained for 3,500 iterations without adding SSPE to the visual tokens. In the second stage, LoRA is introduced, and the model undergoes an additional 1,750 iterations.
The number of trainable parameters in the second stage is approximately 430M, with the newly introduced SVP accounting for only 52M.
Other training and inference settings are consistent with GLaMM~\cite{hanoona2023GLaMM}. Experiments are performed on 4 NVIDIA RTX 3090 GPUs (24GB memory).

\subsection{Comparison to State-of-the-Art}

\begin{table}[htpb]
    \centering
    \caption{
    RIS performance comparison of SVP and existing compressive visual projectors on RefCOCO, RefCOCO+, and RefCOCOg. The best performance is marked in bold.
    }
    \begin{tabular}{lcccccccc}
        \toprule \multirow{2}{*}{Method}                                              & \multicolumn{3}{c}{RefCOCO} & \multicolumn{3}{c}{RefCOCO+} & \multicolumn{2}{c}{RefCOCOg} \\
        \cmidrule(lr){2-4} \cmidrule(lr){5-7} \cmidrule(lr){8-9} & \textit{val}      & \textit{testA}      & \textit{testB}     & \textit{val} & \textit{testA} & \textit{testB} & \textit{val} & \textit{test} \\
        \midrule
        Q-Former{$_{\text{\color{gray}{ICML23}}}$}~\cite{li2023blip2,instructblip} & 71.1 & 72.8 & 67.6 & 55.1 & 59.9 & 50.2 & 60.2  & 61.6 \\  
        Pixel-Shuffle{$_{\text{\color{gray}{CVPR24}}}$}~\cite{chen2024internvl} & 78.0&79.8&75.8&68.8&72.2&65.2&72.4&73.2 \\  
        C-Abstractor{$_{\text{\color{gray}{CVPR24}}}$}~\cite{cha2023honeybee} & 76.3&78.3&72.5&63.2&68.6&57.8&66.4&67.2 \\  
        D-Abstractor{$_{\text{\color{gray}{CVPR24}}}$}~\cite{cha2023honeybee} & 77.3&79.3&73.9&66.3&71.4&61.1&69.4&69.7 \\  
        DeCo{$_{\text{\color{gray}{arXiv24}}}$}~\cite{yao2024deco} & 76.5 & 78.5 & 73.5 & 65.2 & 69.5 & 60.8 & 70.4 & 70.5 \\  
        LDPv2{$_{\text{\color{gray}{arXiv24}}}$}~\cite{chu2023mobilevlm,chu2024mobilevlm} & 78.9&80.9&75.7&69.8&73.4&64.3&72.9&73.7 \\ 
        TokenPacker{$_{\text{\color{gray}{arXiv24}}}$}~\cite{li2024tokenpacker} & 78.7&80.9&76.0&70.0&73.8&64.0&71.3&71.9 \\  
        PVTC{$_{\text{\color{gray}{arXiv25}}}$}~\cite{lu2025internvl} & 76.7 & 79.2 & 73.5 & 65.3 & 70.8 & 59.3 & 69.2 & 69.7 \\  
        \rowcolor{gray!40} SVP & \textbf{81.5} & \textbf{83.1} & \textbf{79.1} & \textbf{75.0} & \textbf{79.1} & \textbf{70.0} & \textbf{76.0} & \textbf{76.3} \\ 
        \bottomrule
    \end{tabular}
    \label{tab:Compress-VS}
\end{table}

\begin{table}[htpb]
    \centering
    \caption{Performance comparison with existing RIS methods under \textbf{cIoU} metric.
    Best performance or the the fewest number of visual tokens is makred in bold, while the second-best performance is underlined. ``Token Num." refers to the average number of visual tokens involved during training and inference per image, which is marked with a ``–" to indicate not applicable for non-MLLM-based methods.
    }
    \begin{tabular}{lccccccccc}
        \toprule \multirow{2}{*}{Method}                                              & \multirow{2}{*}{\makecell{Token \\ Num.}} & \multicolumn{3}{c}{RefCOCO} & \multicolumn{3}{c}{RefCOCO+} & \multicolumn{2}{c}{RefCOCOg} \\
        \cmidrule(lr){3-5} \cmidrule(lr){6-8} \cmidrule(lr){9-10} & & \textit{val}      & \textit{testA}      & \textit{testB}     & \textit{val} & \textit{testA} & \textit{testB} & \textit{val} & \textit{test}   \\
        \midrule
        CRIS{$_{\text{\color{gray}{CVPR22}}}$}~\cite{wang2022cris} & - & 70.5 & 73.2 & 66.1 & 65.3 & 68.1 & 53.7 & 59.9 & 60.4 \\
        LAVT{$_{\text{\color{gray}{CVPR22}}}$}~\cite{yang2022lavt} & - & 72.7 & 75.8 & 68.8 & 62.1 & 68.4 & 55.1 & 61.2 & 62.1 \\
        GRES{$_{\text{\color{gray}{CVPR23}}}$}~\cite{liu2023gres} & - & 73.8 & 76.5 & 70.2 & 66.0 & 71.0 & 57.7 & 65.0 & 66.0 \\
        X-Decoder{$_{\text{\color{gray}{CVPR23}}}$}~\cite{zou2022xdecoder} & - & - & - & - & - & - & - & 64.6 & - \\
        SEEM{$_{\text{\color{gray}{NIPS23}}}$}~\cite{zou2023segment} & - & - & - & - & - & - & - & 65.7 & - \\
        PixelLM{$_{\text{\color{gray}{CVPR24}}}$}~\cite{ren2024pixellm} & 576 & 73.0 & 76.5 & 68.2 & 66.3 & 71.7 & 58.3 & 69.3 & 70.5 \\
        PerceptionGPT-13B{$_{\text{\color{gray}{CVPR24}}}$}~\cite{pi2024perceptiongpt} & 576 & 75.3 & 79.1 & 72.1 & 68.9 & 74.0 & 61.9 & 70.7 & 71.9 \\
        LISA-7B{$_{\text{\color{gray}{CVPR24}}}$}~\cite{lai2023lisa} & 576 & 74.9 & 79.1 & 72.3 & 65.1 & 70.8 & 58.1 & 67.9 & 70.6 \\
        GLaMM{$_{\text{\color{gray}{CVPR24}}}$}~\cite{hanoona2023GLaMM} & 576 & \underline{79.5} & \textbf{83.2} & \underline{76.9} & \textbf{72.6} & \textbf{78.7} & \underline{64.6} & \textbf{74.2} & \textbf{74.9} \\
        F-LMM{$_{\text{\color{gray}{arXiv24}}}$}~\cite{hanoona2023GLaMM} & 576 & 76.1 & - & - & 66.4 & - & - & 70.1 & - \\
        OMG-LLaVA{$_{\text{\color{gray}{arXiv24}}}$}~\cite{OMGLLaVA} & 256 & 78.0 & 80.3 & 74.1 & 69.1 & 73.1 & 63.0 & 72.9 & 72.9 \\
        M$^2$SA{$_{\text{\color{gray}{arXiv25}}}$}~\cite{jang2025mmr} & 256 & 74.6 & 77.6 & 71.0 & 64.0 & 68.1 & 57.6 & 69.0 & 69.3 \\
        \rowcolor{gray!40} SVP & \textbf{$\sim$40} &\textbf{80.2}&\underline{82.9}&\textbf{77.0}&\underline{71.6}&\underline{76.6}&\textbf{65.9}&\underline{73.7}&\underline{74.6}\\
        \bottomrule
    \end{tabular}
    \label{tab:SST-SOTA}
\end{table}

To evaluate the effectiveness of the Semantic Visual Projector (SVP), we replace it with existing compressive visual projectors, keeping an identical experimental setup and ensuring a similar average visual token number across all methods. Tab.~\ref{tab:Compress-VS} presents their performance compared to our SVP.
Consistent with previous findings~\cite{yao2024deco,cha2023honeybee}, Q-Former~\cite{li2023blip2,instructblip}, lacking location priors, produces visual tokens dominated by prominent objects, resulting in poor performance on pixel-level tasks.
The other projectors compared in Tab.~\ref{tab:Compress-VS} are all based on downsampling, exhibiting performance fluctuations within a relatively narrow range, while SVP significantly outperforms all of them.
This gap arises because these downsample-based projectors can be classified as patch-wise visual projectors that use larger patches, which commonly suffer from two issues: (1) larger patch sizes can cause semantic inconsistencies within visual tokens or overlook non-salient semantics, and (2) they cannot dynamically increase the number of visual tokens in complex scenes.
In contrast, SVP generates visual tokens from semantic superpixels rather than patches, effectively avoiding these issues and achieving significantly superior performance.

Table \ref{tab:SST-SOTA} compares state-of-the-art RIS results. 
While it is well known that gIoU is preferred~\cite{liu2023gres,ren2024pixellm,lai2023lisa,jang2025mmr}, the community continues to report cIoU on this benchmark for historical consistency. 
Although this evaluation setting is less favorable for our method, SVP nonetheless achieves competitive performance with markedly fewer visual tokens.

\subsection{Ablation Studies}

\subsubsection{Effectiveness of Main Components}

\setcounter{rownumber}{0}
\begin{table}[t!]
    \centering
    \caption{
    Performance comparison of the GLaMM~\cite{hanoona2023GLaMM}, the proposed method and its variants.
    ``Token Num." denotes the average number of visual tokens per image.
    ``$\mathcal{D}$" denotes replacing the CLIP vision encoder with DINOv2. ``$+\mathcal{P}$" denotes adding SSPE. ``$\mathcal{A}$" denotes the introduction of SSA.
    ``GLaMM + $\mathcal{D}$" denotes replacing the vision encoder of GLaMM with DINOv2, followed by fine-tuning for RIS without SVP.
    }
    \begin{tabular}{c|lccccccccc}
        \toprule
        \multirow{2}{*}{\#} &
        \multirow{2}{*}{Method}              & \multirow{2}{*}{\makecell{Token \\ Num.}}                                & \multicolumn{3}{c}{RefCOCO} & \multicolumn{3}{c}{RefCOCO+} & \multicolumn{2}{c}{RefCOCOg} \\
        \cmidrule(lr){4-6} \cmidrule(lr){7-9} \cmidrule(lr){10-11} & & & \textit{val}      & \textit{testA}      & \textit{testB}     & \textit{val} & \textit{testA} & \textit{testB} & \textit{val} & \textit{test} \\
        \midrule
        \rownumber & GLaMM~\cite{hanoona2023GLaMM} & 576 & 80.8 & 82.8 & 77.8 & 74.3 & 78.9 & 68.3 & 75.3 & 76.0 \\ 
        \rownumber & GLaMM~\cite{hanoona2023GLaMM} + $\mathcal{D}$ & 576 & 81.7 & 83.1 & 78.2 & 74.9 & 79.4 & 70.1 & 76.5 & 76.9  \\ 
        \rownumber & Baseline & $\sim$40 & 71.6 & 75.2 & 69.0 & 56.4 & 61.1 & 51.2 & 61.7 & 62.9 \\ 
        \rownumber & Baseline + $\mathcal{D}$ & $\sim$40 & 77.0 & 78.7 & 72.7 & 65.4 & 71.7 & 59.2 & 69.4 & 69.7 \\ 
        \rownumber & Baseline + $\mathcal{D}$ + $\mathcal{P}$ & $\sim$40 & 80.1 & 81.7 & 76.9 & 72.3 & 75.5 & 67.1 & 73.8 & 74.6 \\ 
        \rownumber & Baseline + $\mathcal{D}$ + $\mathcal{P}$ + $\mathcal{A}$ & $\sim$40 & 81.5 & 83.1 & 79.1 & 75.0 & 79.1 & 70.0 & 76.0 & 76.3 \\ 
        \bottomrule
    \end{tabular}
    \label{tab:SST-Abl}
\end{table}

Tab.~\ref{tab:SST-Abl} validates the effectiveness of main components within the SVP.
The baseline represents a simplified implementation, where coarse semantic superpixel embeddings yielded by average pooling are directly projected into visual tokens via an MLP without any additional positional embedding.
However, CLIP's image embeddings lack fine-grained semantics, primarily capturing only the most salient objects in the image. This limitation is exacerbated by average pooling, leading to performance that is significantly inferior to GLaMM~\cite{hanoona2023GLaMM}.
In contrast, DINOv2~\cite{dinov2} explicitly models semantics at multiple levels across all image positions.
As shown in the fourth row of Tab.~\ref{tab:SST-Abl}, compared to the baseline, replacing the vision encoder with DINOv2 leads to a substantial performance improvement.
Although Sec.\ref{sec:embedding_fusion} attempts to fuse embeddings from multiple layers or different vision encoders, the results indicate that DINOv2 alone is more effective and simpler.

The second row of Tab.~\ref{tab:SST-Abl} replaces GLaMM's vision encoder with DINOv2 and fine-tunes it again for RIS.
The results indicate that incorporating DINOv2 is also effective for GLaMM but does not benefit as significantly as our SVP.
This reinforces the notion that replacing the vision encoder effectively addresses the baseline's unique challenges, mitigating information loss during the computation of semantic superpixel embeddings.

The fifth row of Tab.~\ref{tab:SST-Abl} introduces Semantic Superpixel Positional Embedding (SSPE) to explicitly enhance the visual tokens' awareness of the geometry and position of the semantic superpixels, significantly improving performance.
Notably, RefCOCO+, which disables absolute positioning, often relies on contextual relationships, relative positions, and relative sizes to differentiate similar objects, leading to the most pronounced boost from SSPE.
Interestingly, we find that SVP can still ground some references related to size or position even without SSPE, particularly concerning absolute positioning, rather than failing entirely.
This may be attributed to the absolute positional embeddings added by the vision encoder, which offer some positional cues to the MLLM.

The sixth row of Tab.~\ref{tab:SST-Abl} introduces the Semantic Superpixel Aggregator (SSA) to refine coarse semantic superpixel embeddings by adaptively aggregating semantics both within and outside the superpixels, further enhancing performance.
Similarly, the benefits of SSA are most pronounced on RefCOCO+, where fine-grained attributes and object relationships are crucial.

The sixth row is also the final version of SVP. Despite requiring only $\sim$6.9\% of the visual token budget used by GLaMM, SVP surpasses the original GLaMM and achieves performance rivaling GLaMM with DINOv2, demonstrating substantial efficiency gains with virtually no performance compromise, even on the challenging RIS task.

\subsubsection{Effectiveness of SSPE Encoder Designs}

\begin{table}[htpb]
    \centering
    \captionof{table}{
    Performance comparison of the SSPE encoder and its variants.
    ``BBox-MLP" denotes adopting a MLP to project the bounding box of the semantic superpixel into SSPE.
    ``Mask-MLP" denotes projecting the resized mask of the superpixel into SSPE.
    ``Mask-Att." is the proposed SSPE encoder.
    }
    \begin{tabular}{lcccccccc}
        \toprule \multirow{2}{*}{Method}                                              & \multicolumn{3}{c}{RefCOCO} & \multicolumn{3}{c}{RefCOCO+} & \multicolumn{2}{c}{RefCOCOg} \\
        \cmidrule(lr){2-4} \cmidrule(lr){5-7} \cmidrule(lr){8-9} & \textit{val}      & \textit{testA}      & \textit{testB}     & \textit{val} & \textit{testA} & \textit{testB} & \textit{val} & \textit{test} \\
        \midrule
        w/o SSPE & 77.0 & 78.7 & 72.7 & 65.4 & 71.7 & 59.2 & 69.4 & 69.7 \\ 
        BBox-MLP & 78.4 & 80.4 & 75.3 & 69.1 & 74.0 & 63.5 & 70.9 & 72.6 \\ 
        Mask-MLP & 77.7 & 79.3 & 74.4 & 67.6 & 72.8 & 61.2 & 70.7 & 71.0 \\ 
        Mask-Att. & 80.1 & 81.7 & 76.9 & 72.3 & 75.5 & 67.1 & 73.8 & 74.6 \\ 
        BBox-MLP + Mask-Att. & 78.1 & 80.6 & 74.9 & 68.6 & 74.1 & 64.0 & 70.9 & 72.0 \\ 
        \bottomrule
    \end{tabular}
    \label{tab:SSPE-Abl}
\end{table}

\begin{figure}[htpb]
    \centering
    \includegraphics[width=0.975\linewidth]{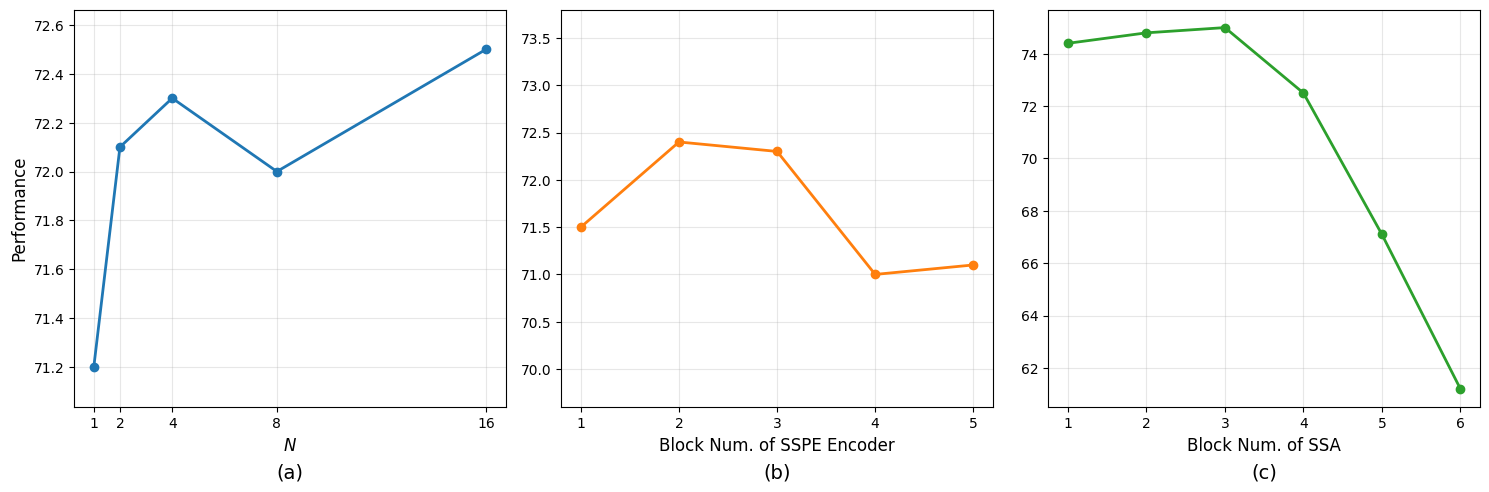}
    \captionof{figure}{
        Parameter tuning results for the number of queries $N$, the number of blocks for the SSPE encoder, and the SSA.
        ``Performance" is evaluated on RefCOCO+ \textit{val}.
        }
    \label{fig:N-ssa-sspe-tuning}
    \Description{}
\end{figure}

\begin{figure}[htpb]
    \centering
    \includegraphics[width=0.975\linewidth]{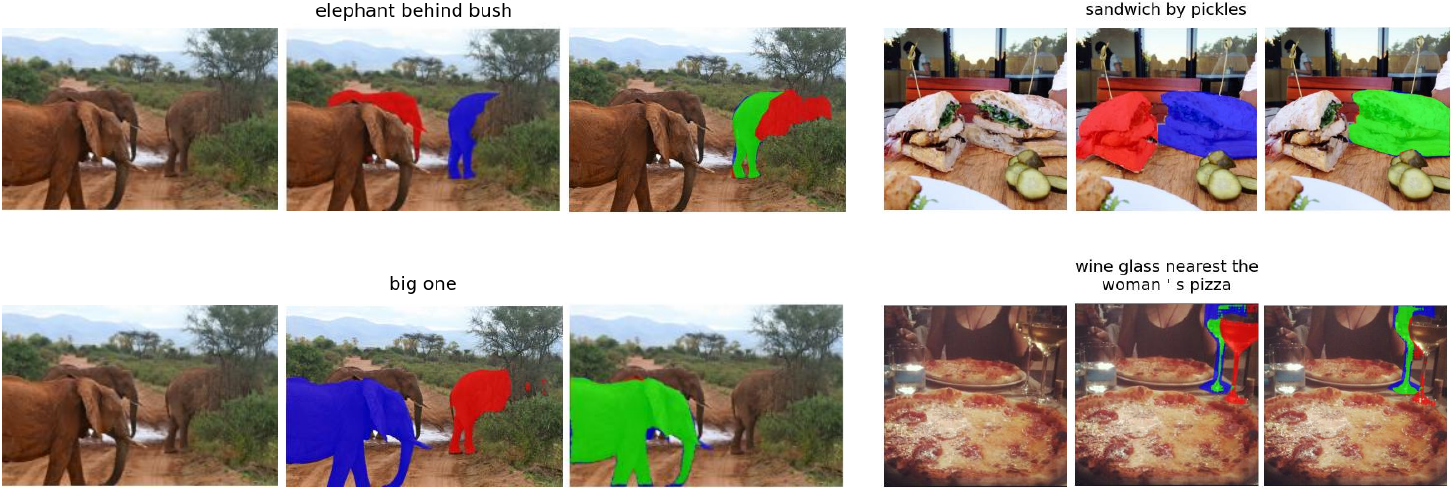}
    \caption{
        Qualitative comparison of the RIS results with and without SSPE.
        Each row presents the same sample: the original image, the RIS result without SSPE (Tab.~\ref{tab:SST-Abl}, row 4) and the result with SSPE (Tab.~\ref{tab:SST-Abl}, row 5).
        The referring expression is annotated above.
        Green masks indicate true positives (TP), red masks denote false positives (FP), and blue masks represent false negatives (FN).
        }
    \label{fig:sspe_viz}
    \Description{}
\end{figure}

Tab.~\ref{tab:SSPE-Abl} explores several different methods for computing  Semantic Superpixel Positional Embedding (SSPE).
Projecting bounding boxes of semantic superpixels for SSPE also yields improvements, indicating that even coarse-grained size or positional information enhances the MLLM's understanding of image structure.
To provide finer geometric and positional cues, we computed SSPE using semantic superpixel masks.
Resizing the superpixel masks to a fixed size and projecting them into SSPE yielded unsatisfactory results, likely due to the MLP's challenges in processing spatial information conveyed by binary masks.
In contrast, our SSPE encoder, which utilizes cross-attention with multiple learnable queries to capture the spatial details, demonstrates superior performance.
Moreover, combining the outputs from both the bounding box projections and the SSPE encoder degrades performance to the level of using box projections alone. This may be because the model prioritizes the easier-to-learn box projections, leading to a local optimum.

Fig.~\ref{fig:N-ssa-sspe-tuning} (a) illustrates the effect of the number of learnable queries $N$ in the SSPE encoder. Increasing
$N$ beyond $1$ generally enhances performance. We choose $N = 4$ to balance performance and computational efficiency.

Fig.~\ref{fig:sspe_viz} compares the RIS results with and without SSPE, illustrating that image structural cues from SSPE effectively correct RIS errors regarding object relationships, relative sizes, and relative positions.

\subsubsection{Effectiveness of SSA Designs}

\begin{table}[htpb]
    \centering
    \caption{
    Performance comparison of the SSA and its variants.
    ``w/o sharing" denotes not sharing SSPE at the superpixel level.
    ``Att. Bias" denotes contraining the attention area within the superpixel by adding attention bias.
    ``w/ Sharing" is the proposed SSA.
    }
    \begin{tabular}{lcccccccc}
        \toprule \multirow{2}{*}{Method}                                              & \multicolumn{3}{c}{RefCOCO} & \multicolumn{3}{c}{RefCOCO+} & \multicolumn{2}{c}{RefCOCOg} \\
        \cmidrule(lr){2-4} \cmidrule(lr){5-7} \cmidrule(lr){8-9} & \textit{val}      & \textit{testA}      & \textit{testB}     & \textit{val} & \textit{testA} & \textit{testB} & \textit{val} & \textit{test} \\
        \midrule
        w/o SSA & 80.1 & 81.7 & 76.9 & 72.3 & 75.5 & 67.1 & 73.8 & 74.6 \\ 
        w/o Sharing & 78.7 & 80.9 & 75.7 & 69.2 & 74.4 & 62.9 & 71.2 & 71.6 \\ 
        w/o Sharing + Att. Bias & 80.8 & 82.1 & 77.2 & 73.4 & 77.2 & 67.0 & 74.7 & 75.5 \\ 
        w/ Sharing & 81.5 & 83.1 & 79.1 & 75.0 & 79.1 & 70.0 & 76.0 & 76.3 \\ 
        \bottomrule
    \end{tabular}
    \label{tab:SSA-Abl}
\end{table}

Tab.~\ref{tab:SSA-Abl} validates the effectiveness of sharing SSPE at the superpixel level within the Semantic Superpixel Aggregator (SSA).
The second row shows that without incorporating location priors for semantic superpixels, fully free-form cross-attention tends to neglect non-salient regions and local details, leading to decreased performance.
By introducing attention biases that constrain the attention within the semantic superpixels, the model captures local details more effectively, resulting in moderate performance improvements.
Finally, SSA achieves optimal performance through superpixel-level SSPE sharing, balancing local details with contextual information essential for RIS, and shows significant enhancements on RefCOCO+, which relies heavily on contextual relationships. 

\begin{figure}[htpb]
    \centering
    \includegraphics[width=0.925\linewidth]{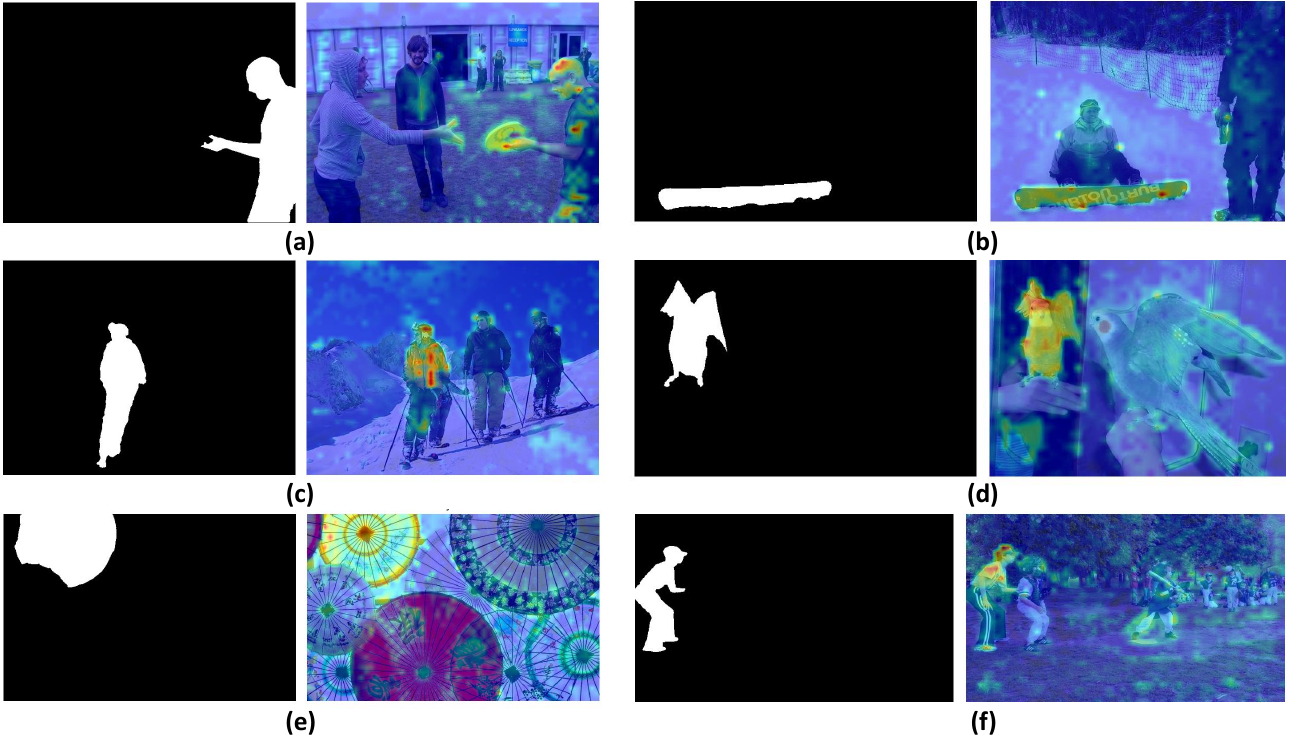}
    \caption{
        Cross-attention maps of queries to the patch-wise image embeddings in SSA.
        Each row presents the same sample: the semantic superpixel corresponding to the query and the cross-attention map (overlaid on the original image).
        }
    \label{fig:ssa_att_map}
    \Description{}
\end{figure}

The cross-attention maps of SSA in Fig.~\ref{fig:ssa_att_map} further validate our design.
These maps reveal several key behaviors.
First, the query consistently prioritizes the semantics within its corresponding semantic superpixel, indicating that sharing the SSPE at the superpixel level imparts a local inductive bias.
Second, attention within a superpixel is uneven but concentrates on discriminative regions, such as a bird's head, brightly colored clothing, or the circular shape of an umbrella.
Thus, SSA-refined superpixel embeddings effectively resample salient details within the superpixel, better capturing fine-grained attributes relevant to RIS.
Moreover, SSA's attention maps can also capture context, including closely interacting objects (e.g., a frisbee in a person's hand, or an athlete on a snowboard, as shown in Fig.~\ref{fig:ssa_att_map}~a-b) and objects of the same category (Fig.~\ref{fig:ssa_att_map}~c-f).
For same-category objects, attention tends to focus on the edges—a behavior that may help learn relative positions and sizes while preserving semantic distinctness.
Finally, some attention is diffusely distributed across the background, likely to capture the overall scene context.

\subsubsection{Parameter Tuning}
Fig.~\ref{fig:N-ssa-sspe-tuning}~(b) and (c) show the effect of varying the number of blocks in the SSPE encoder and SSA.
Performance peaks with 2-3 blocks before declining, possibly due to over-parameterization hindering learning in the SSPE encoder or SSA.
We set the number of blocks to 3 for both modules to reduce hyperparameter complexity.

\subsubsection{Qualitative Analysis}

\begin{figure}[t!]
    \centering
    \includegraphics[width=0.975\linewidth]{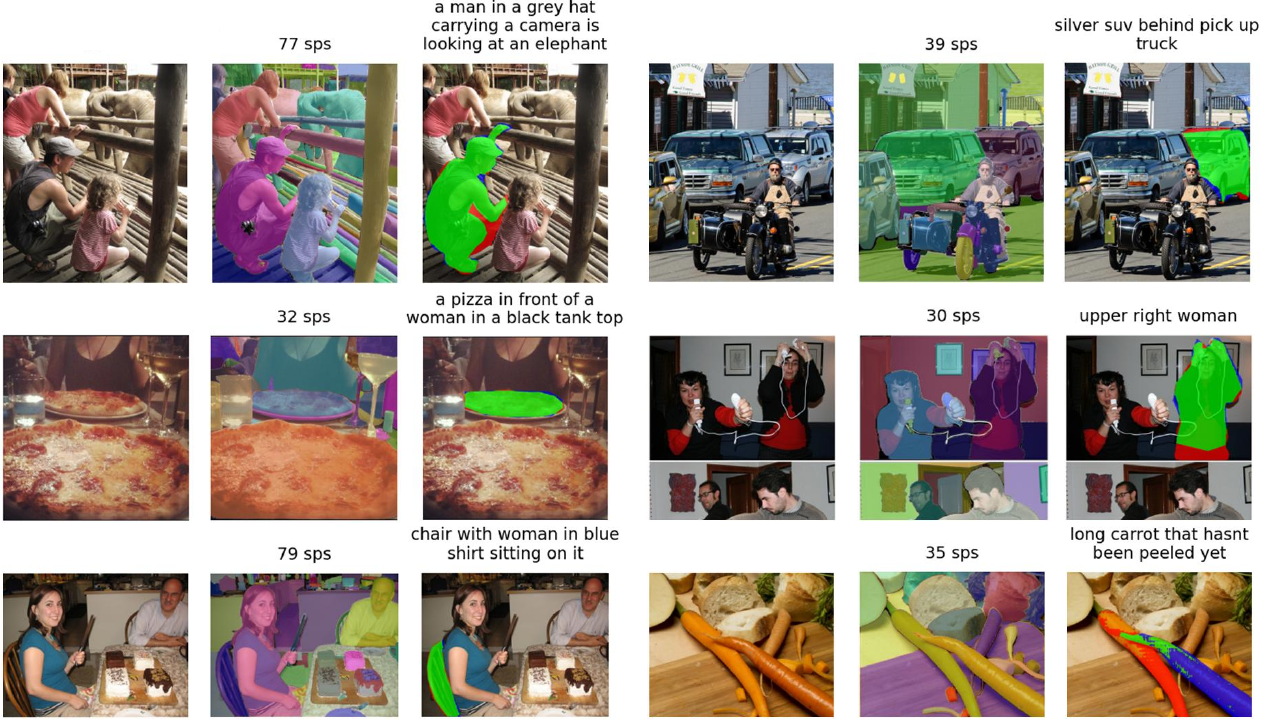}
    \caption{
    Typical RIS results obtained with our method.
    Each row presents the same sample: the original image, the image annotated with semantic superpixels (number of superpixels noted above), and the RIS result (with the referring expression noted above). The bottom right corner shows a failure example.
    }
    \label{fig:main_viz}
    \Description{}
\end{figure}

\begin{figure}[htpb]
    \centering
    \includegraphics[width=0.975\linewidth]{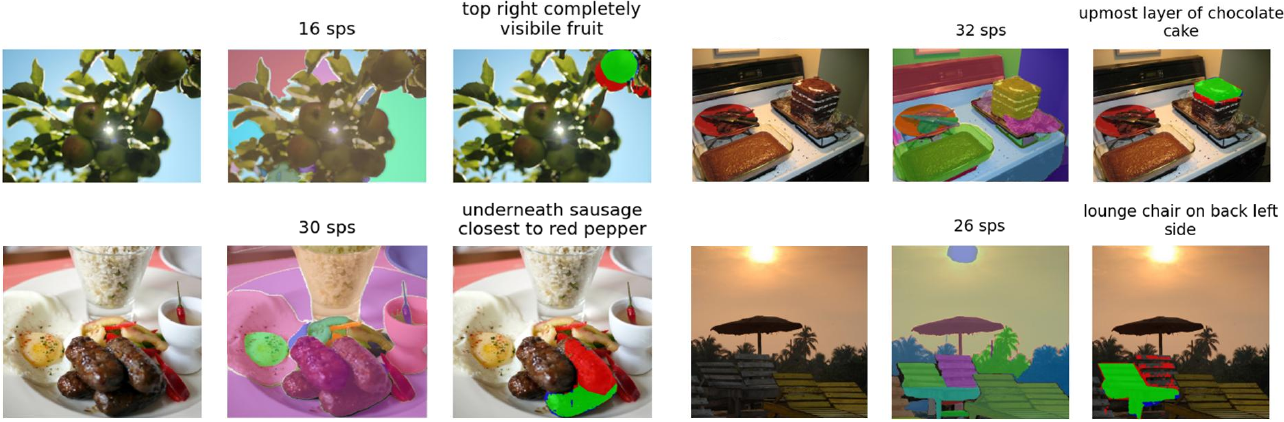}
    \caption{
    The first three sets demonstrate target objects that are under-segmented, while the bottom right example shows an over-segmented target. The arrangement aligns with Fig.~\ref{fig:main_viz}.
    }
    \label{fig:over_under_seg}
    \Description{}
\end{figure}

Fig.~\ref{fig:main_viz} showcases typical results from the SVP.
In most cases, semantic superpixels successfully separate the target object from the environment. This enables the MLLM to select the correct visual token from multiple candidates that match the referring expression, significantly reducing the difficulty of segmentation.
The lower right corner of Fig.~\ref{fig:main_viz} shows a failure case where SVP struggles with elongated objects. Their extensive boundaries with the background complicate the aggregation of their pure semantics.
More visualization results can be found in Sec.~\ref{sec:more_viz}.

Fig.~\ref{fig:over_under_seg} examines the MLLM's ability to accurately segment target objects when semantic superpixels are either under-segmented (significantly larger than the target) or over-segmented (divided into multiple superpixels).
The experiments show that MLLMs do not simply match visual tokens to text and select the superpixel that best aligns with the referring expression.
Instead, in cases of under-segmentation, the MLLM can still analyze the spatial structure within the semantic superpixel to locate the target object. During over-segmentation, it effectively combines multiple visual tokens to fully reconstruct the target.
These findings validate the generalization capability of SVP, demonstrating that its performance is not solely reliant on SAM's segmentation abilities.
Even with suboptimal division of semantic superpixels, the MLLM retains a certain degree of corrective capacity.
However, under-segmented semantic superpixels are more likely to yield higher false positives, as reconstructing the internal structure from a single visual token can be challenging.
Additionally, because semantic superpixels tend to select higher-level masks, under-segmentation is more common, leading to relatively poor performance on the cIoU metric, which is more sensitive to false positives.

\subsection{Efficiency Analysis}

\begin{table}[htpb]
    \centering
    \caption{
    Comparison of our method to GLaMM~\cite{hanoona2023GLaMM} with DINOv2 vision encoder (denoted as GLaMM + $\mathcal{D}$) across visual token number, training efficiency, inference efficiency, and RIS performance.
    ``Token Num." denotes the average number of visual tokens involved during training and inference per image.
    Training time is the average iteration time per data batch, while inference time is the average processing time per image.
    ``Max Mem." denotes the maximum GPU memory usage during training or inference.
    ``Performance" is averaged on all validation and test sets.
    Better results are marked in bold.
    }
    \label{tab:ST-Efficiency}
    \begin{tabular}{lcccccc}
    \hline
    \multirow{2}{*}{Method} & \multirow{2}{*}{Token Num.} & \multicolumn{2}{c}{Training} & \multicolumn{2}{c}{Inference} & \multirow{2}{*}{Performance} \\
    \cmidrule(r){3-4} \cmidrule(r){5-6}
    & & \makecell{Time \\ (s)} & \makecell{Max Mem. \\ (GB)} & \makecell{Time \\ (ms)} & \makecell{Max Mem. \\ (GB)} &  \\
    \hline
    GLaMM + $\mathcal{D}$ & 576 & 46.0 & 21.4
    & 384 & 19.9 & \textbf{77.6} \\
    \rowcolor{gray!40} SVP & \textbf{$\sim$40} & \textbf{12.4} & \textbf{21.1}
        & \textbf{152} & \textbf{15.0} & 77.5 \\
    \hline
    \end{tabular}
\end{table}

\begin{figure}[htpb]
    \centering
    \includegraphics[width=0.925\linewidth]{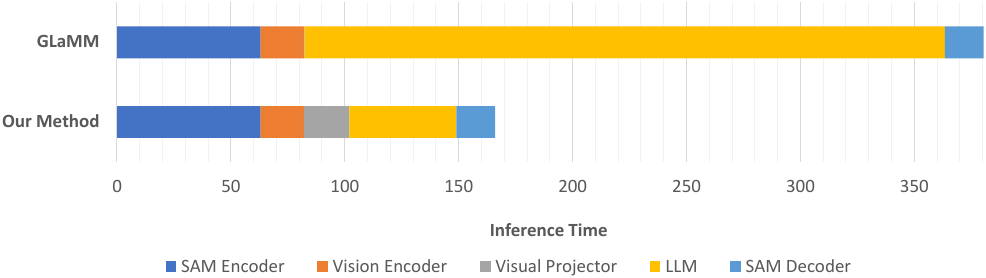}
    \caption{
        A component-wise breakdown of the inference time for GLaMM and our method.
        The total time may differ slightly from the results presented in Tab.~\ref{tab:ST-Efficiency} due to minor differences in the testing environment and inherent random fluctuations.
        }
    \label{fig:time_decomp}
    \Description{}
\end{figure}

Tab.~\ref{tab:ST-Efficiency} compares the efficiency of GLaMM~\cite{hanoona2023GLaMM} with our method.
The results show that SVP reduces the average number of visual tokens by 93.1\%, while training and inference times decrease by 60\% to 70\%. Although memory reductions during training are minimal due to various optimization techniques, inference memory usage drops by approximately 25\%.
Additionally, while training and inference speed are roughly tripled, SVP’s performance is just 0.1\% lower than that of GLaMM using DINOv2~\cite{dinov2} as the vision encoder.

It is important to note that actual computation time does not fully align with theoretical complexity, which has also been reported in previous research~\cite{cha2023honeybee}.
Given that training and inference speed can vary based on hardware conditions, the number of visual tokens serves as a more stable and comparable metric for evaluating visual token reduction methods.

Fig.~\ref{fig:time_decomp} decomposes the inference times for GLaMM and our method, revealing that the LLM is the primary efficiency bottleneck in existing MLLM-based RIS frameworks.
Our SVP addresses this by reducing the number of visual tokens sent to the LLM, thereby decreasing its latency by approximately 234 ms.
In contrast, the SVP itself introduces a minimal overhead of only 20 ms, of which 17 ms are attributed to SAM decoder, with the rest originating from the SSPE Encoder, SSA, and MLP.
Consequently, owing to the high efficiency of SVP’s components, the latency reduction from SVP substantially outweighs its computational cost, yielding a significant improvement in overall efficiency.

\section{Conclusion}

We find that traditional patch-wise visual projectors consistently suffer from one of two issues: visual token redundancy or ambiguous token semantics.
To address both problems, we propose a Semantic Visual Projector (SVP), which computes visual tokens from fundamental visual elements identified by SAM and dynamically adjusts the token number according to scene complexity.
We further identify that the default positional embedding struggles to represent the image structure formed by semantic superpixels.
We resolve this by integrating Semantic Superpixel Positional Embedding (SSPE), generated by an adaptive encoder that captures superpixel geometry and position through multiple queries.
Finally, we present a Semantic Superpixel Aggregator (SSA), which enhances local detail awareness inside superpixels through a cross-attention layer sharing SSPE at the superpixel level, while also capturing essential surrounding context for RIS.
Our method significantly boosts training and inference efficiency for MLLMs on RIS while maintaining performance, and it greatly surpasses existing compressive visual projectors.

\section{Future Work}

Due to limited computational resources, we did not evaluate SVP on a broader range of vision or multimodal tasks.
For instance, conducting a fair comparison on a comprehensive multimodal benchmark necessitates training the entire LLM instead of using parameter-efficient fine-tuning and replicating an LLaVA-like~\cite{liu2023llava,liu2023improvedllava} training schedule, including pre-training and instruction tuning~\cite{cha2023honeybee,yao2024deco,lu2025internvl}, which exceeds our current computational capabilities.

Given limited computing resources, we choose to focus on RIS for the following reasons:
(1) Representativeness: RIS is one of the most challenging multimodal tasks, requiring detailed text parsing and pixel-level grounding, making it an ideal testbed for validating the model's multimodal understanding after applying SVP.
(2) Novelty: Utilizing MLLMs to ground pixels represents a relatively new research direction, with limited studies addressing their efficiency.
(3) Effectiveness: SVP can reuse SAM within MLLM-based RIS frameworks, maximizing their potential.

Theoretically, SVP can be generalized effectively to other multimodal tasks (e.g., visual question answering) while still enhancing the overall efficiency of MLLM.
Our experimental results on RIS demonstrate that SVP-equipped MLLMs effectively recognize object categories and attributes and capture their spatial and interactive relationships.
The rich semantic and structural information in SVP's visual tokens suggests its applicability to a wider range of MLLM scenarios.
Furthermore, as shown in Fig.~\ref{fig:time_decomp}, when compared to a standard MLLM without SAM (e.g., GLaMM without SAM), the additional overhead introduced by SVP (83ms, which includes the runtime for both the SAM encoder and SVP itself) is still substantially less than the computational savings it yields for the LLM (234ms).
Thus, even for general multimodal tasks, adding SVP to MLLMs that do not natively rely on external tools offers a significant efficiency improvement.
In future work, we aim to assess the effectiveness and efficiency of SVP across a wider variety of MLLMs and tasks, including classical multimodal applications such as VQA and image captioning, as well as downstream tasks like semantic segmentation and image generation.

Finally, due to the larger region areas corresponding to visual tokens, compressive visual projectors (including SVP) tend to produce more false positives for RIS, which in turn degrades cIoU.
For SVP, one potential remedy is to generate a greater number of finer-grained semantic superpixels, e.g., by computing semantic superpixels within local sliding windows across the image. However, this would introduce additional computational overhead, necessitating a trade-off in practical applications.

\begin{acks}
This research was funded by Zhejiang Province Pioneer Research and Development Project "Research on Multi-modal Traffic Accident Holographic Restoration and Scene Database Construction Based on Vehicle-cloud Intersection" (Grant No. 2024C01017).
\end{acks}

\bibliographystyle{ACM-Reference-Format}
\bibliography{ref}

\clearpage

\appendix

\section{Necessity of Replacing Vision Encoder}
\label{sec:replace_visual_encoder}

\begin{figure}[htpb]
    \centering
    \includegraphics[width=0.975\linewidth]{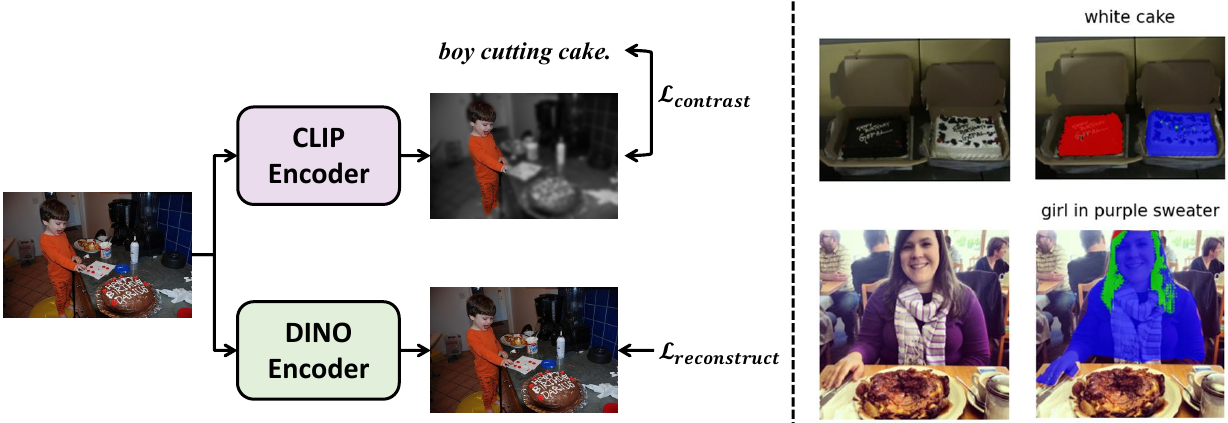}
    \caption{
        Motivation for replacing the vision encoder. Left: The text-image alignment pretraining of CLIP often overlooks attribute information in non-salient areas of images, such as color. In contrast, self-supervised representation learning models like DINOv2 effectively capture details from all regions by leveraging reconstruction loss. Right: Failure case of the baseline.
    }
    \label{fig:Sec5/DINO_better}
    \Description{}

    \centering
    \includegraphics[width=0.975\linewidth]{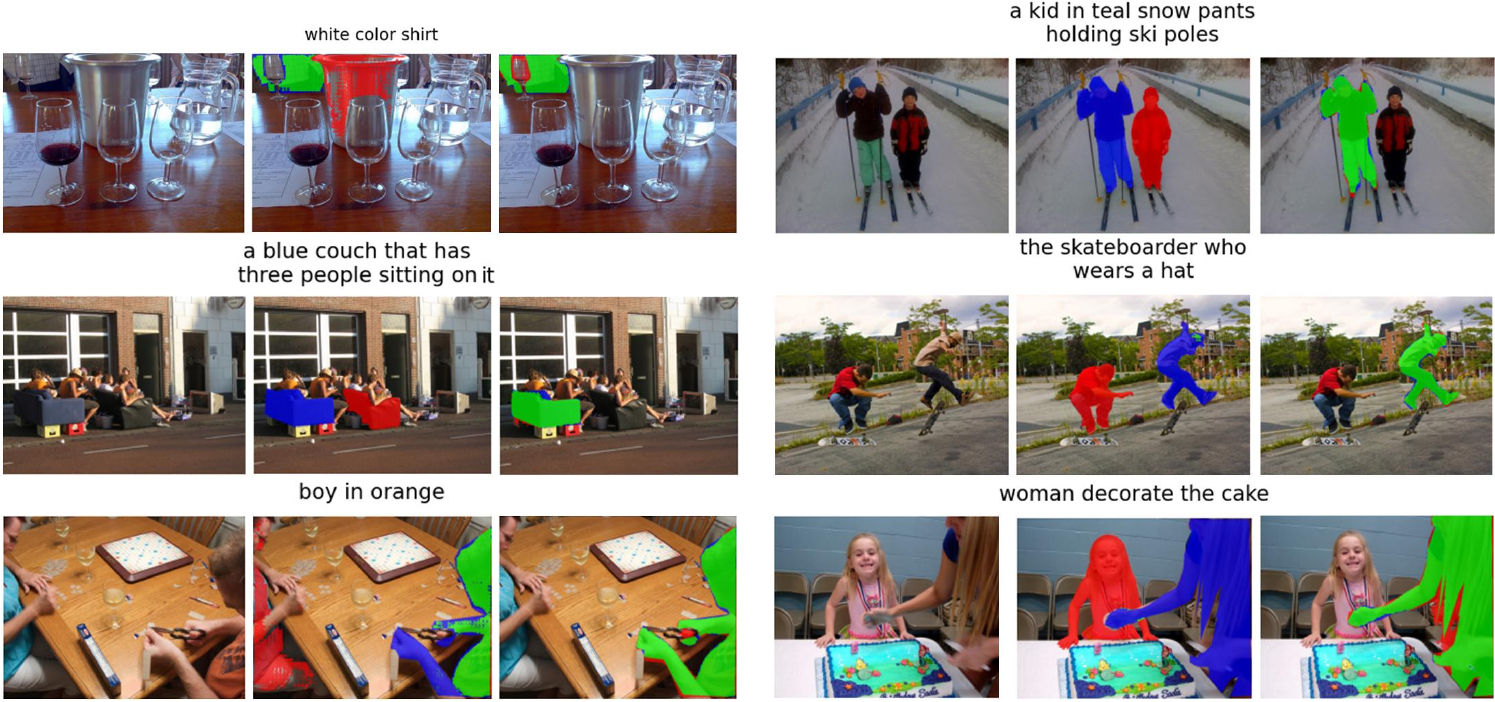}
    \caption{
        Qualitative comparison of the RIS results using CLIP and DINOv2 as vision encoders.
        Each row presents the same sample: the original image, the RIS result from the baseline (Tab.~\ref{tab:SST-Abl}, row 3), and the result after replacing the vision encoder with DINOv2 (Tab.~\ref{tab:SST-Abl}, row 4).
        The referring expression is annotated above.
        }
    \label{fig:CLIPvsDINO}
    \Description{}
\end{figure}

Visual analysis reveals that the baseline's visual tokens lack the fine-grained semantics essential for RIS, causing confusion in distinguishing target objects based on attributes.
As shown in Fig.~\ref{fig:Sec5/DINO_better} and Fig.~\ref{fig:CLIPvsDINO}, the baseline often confuses basic color attributes, struggling to differentiate clearly contrasted colors such as white and black, and fails to identify more complex attributes like actions or decorations.
This inability to distinguish attributes leads the baseline to overly depend on category information during segmentation.
For example, in the bottom right corner of Fig.~\ref{fig:Sec5/DINO_better}, it only segments the long hair of the ``girl" while deliberately ignoring the color description ``purple sweater."
In some examples in Fig.~\ref{fig:CLIPvsDINO}, the baseline attempts to recall targets by segmenting all objects of the same class mentioned in the referring expression, resulting in significant false positives.

The baseline's visual tokens lack fine-grained semantics primarily because the CLIP visual encoder focuses on category-related semantics while neglecting non-salient objects.
Previous studies have suggested that the text used for CLIP's image-text alignment pre-training lacks descriptions of object attributes and mainly depicts primary objects~\cite{wu2022well,baron2024real,roth2023waffling,zhao2022explainable,zhao2022vl}, thereby limiting the representational ability of image embeddings.
This issue is not critical for the MLP visual projector, as the visual encoder is generally over-parameterized (for instance, the inner dimension of CLIP-L is $1024$, which exceeds the input patch dimension of $588$). Therefore, patch-wise image embeddings implicitly retain original image information available for further analysis by MLLMs.
However, such subtle information can easily be disrupted by the average pooling during the computation of coarse semantic superpixel embeddings, ultimately preventing the baseline model from capturing semantic details.

In contrast, self-supervised representation learning models such as DINOv2~\cite{dinov2} achieve patch-level image reconstruction by capturing local details throughout the image.
Consequently, the visual tokens produced by compressing DINOv2 image embeddings retain rich fine-grained semantics. As shown in Fig.~\ref{fig:CLIPvsDINO}, replacing CLIP-L with DINOv2 significantly improves the recognition capability for various attributes.

\section{Effectiveness of Embedding Fusion}
\label{sec:embedding_fusion}

\begin{table}[htpb]
    \centering
    \caption{
    Performance comparison of multi-layer embedding fusion and cross-vision encoder fusion with SVP.
    ``$\mathcal{C}(3, 12, 22)$" denotes fusing semantic superpixel embeddings from the outputs of the 3rd, 12th, and 22nd layers of CLIP.
    ``$\mathcal{D}$" denotes replacing the CLIP vision encoder with DINOv2.
    ``$\mathcal{C}(22)$" and ``$\mathcal{D}(23)$" correspond to the baseline and the final design of SVP.
    }
    \begin{tabular}{lcccccccc}
        \toprule \multirow{2}{*}{Method}                                              & \multicolumn{3}{c}{RefCOCO} & \multicolumn{3}{c}{RefCOCO+} & \multicolumn{2}{c}{RefCOCOg} \\
        \cmidrule(lr){2-4} \cmidrule(lr){5-7} \cmidrule(lr){8-9} & \textit{val}      & \textit{testA}      & \textit{testB}     & \textit{val} & \textit{testA} & \textit{testB} & \textit{val} & \textit{test} \\
        \midrule
        $\mathcal{C}(22)$ & 71.6 & 75.2 & 69.0 & 56.4 & 61.1 & 51.2 & 61.7 & 62.9 \\ 
        $\mathcal{C}(3, 12, 22)$ & 69.9 & 72.0 & 68.2 & 52.5 & 57.4 & 48.5 & 59.4 & 60.2 \\ 
        $\mathcal{D}(23)$ & 81.5 & 83.1 & 79.1 & 75.0 & 79.1 & 70.0 & 76.0 & 76.3 \\ 
        $\mathcal{D}(19, 21, 23)$ & 81.0 & 82.7 & 78.9 & 74.5 & 78.3 & 69.8 & 75.6 & 76.2 \\ 
        $\mathcal{D}(23) + \mathcal{C}(22)$ & 79.2 & 80.7 & 76.4 & 70.1 & 74.3 & 63.4 & 71.7 & 72.1 \\ 
        \bottomrule
    \end{tabular}
    \label{tab:DINO-CLIP-Abl}
\end{table}

Prior works~\cite{jiang2023clip,tong2024cambrian1} suggest that integrating multi-layer embeddings from vision encoders or combining outputs from different encoders can enhance MLLM performance.
This section also aims to explore the use of shallow embeddings from CLIP to address the lack of low-level semantics, such as color and texture attributes, in its deep embeddings.
Following COMM~\cite{jiang2023clip}, we apply SVP to embeddings from various layers or vision encoders and linearly combine the outputs of multiple SVPs using learnable weights. The results are presented in Tab.~\ref{tab:DINO-CLIP-Abl}.

Compared to the baseline, directly integrating the inadequately processed shallow embeddings from CLIP degrades performance.
Additionally, as COMM~\cite{jiang2023clip} noted that shallow embeddings from DINOv2 are detrimental to MLLMs, we only attempt to merge the last few layers of DINOv2, but this also does not yield improvements.
Finally, combining outputs from CLIP and DINOv2 results in performance that falls between the individual performances, indicating that CLIP is unable to provide more visual information for RIS to DINOv2.

\section{Comparison with Alternative Superpixel Generation Methods}
\label{sec:slic_vs_sam}

\begin{figure}[t!]
    \centering
    \includegraphics[width=0.975\linewidth]{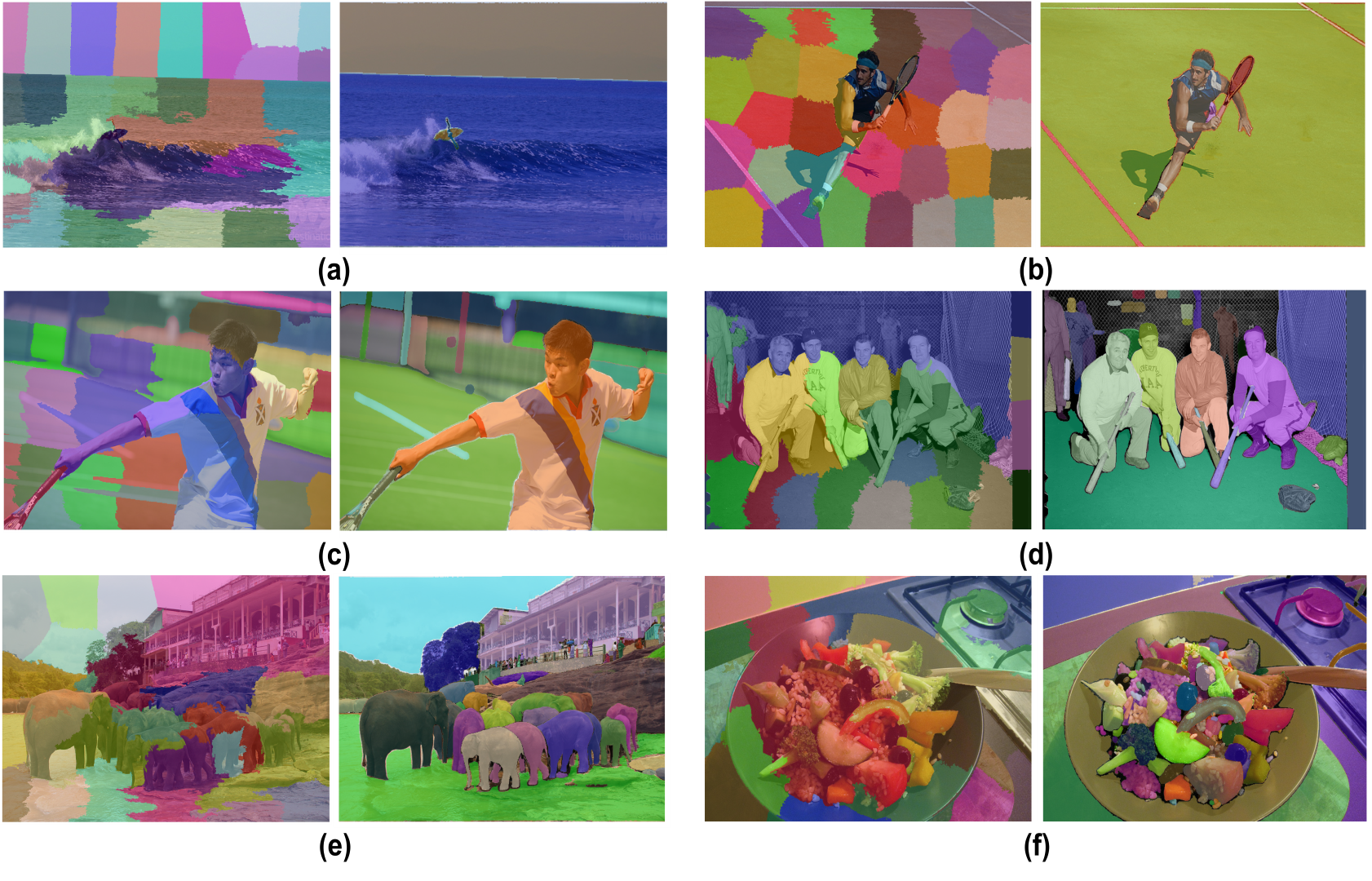}
    \captionof{figure}{
        Comparison of traditional superpixel generated by SLIC~\cite{achanta2012slic} and semantic superpixel generated by SAM~\cite{Kirillov2023SAM}.
        }
    \label{fig:slic_vs_sam}
    \Description{}
\end{figure}

\begin{table}[htpb]
    \centering
    \captionof{table}{
    Performance comparison of the SVP with traditional superpixels generated by SLIC~\cite{achanta2012slic} and semantic superpixels generated by MobileSAM\cite{MobileSAM} or SAM\cite{Kirillov2023SAM}.
    }
    \resizebox{1.0\textwidth}{!}{
    \begin{tabular}{lccccccccc}
        \toprule
        \multirow{2}{*}{Method}              & \multirow{2}{*}{\makecell{Token \\ Num.}}                                & \multicolumn{3}{c}{RefCOCO} & \multicolumn{3}{c}{RefCOCO+} & \multicolumn{2}{c}{RefCOCOg} \\
        \cmidrule(lr){3-5} \cmidrule(lr){6-8} \cmidrule(lr){9-10} & & \textit{val}      & \textit{testA}      & \textit{testB}     & \textit{val} & \textit{testA} & \textit{testB} & \textit{val} & \textit{test} \\
        \midrule
        SLIC Superpixel & $\sim$40 & 75.9 & 78.3 & 73.7 & 65.8 & 69.7 & 60.8 & 70.0 & 70.2 \\ 
        Semantic Superpixel w/ MobileSAM & $\sim$40 & 80.0 & 81.8 & 77.5 & 73.3 & 77.1 & 68.3 & 74.1 & 74.6 \\ 
        Semantic Superpixel w/ SAM & $\sim$40 & 81.5 & 83.1 & 79.1 & 75.0 & 79.1 & 70.0 & 76.0 & 76.3 \\ 
        \bottomrule
    \end{tabular}
    }
    \Description{}
    \label{tab:SLIC-SAM}
\end{table}

Fig.~\ref{fig:slic_vs_sam} compares superpixels generated by the traditional SLIC algorithm~\cite{achanta2012slic} and the semantic superpixels from SAM~\cite{Kirillov2023SAM}.
Since SLIC superpixels are only based on low-level features such as position and color, they often merge foreground and background into the same superpixel, disrupting semantic consistency.
Large objects can easily be divided into multiple superpixels, damaging semantic coherence.
Multiple crowded objects tend to mix together, diminishing the distinguishability of target objects.
Moreover, SLIC lacks the ability to adjust the number of superpixels based on scene complexity, leading to redundancy in simple scenes and insufficient detail in complex scenarios.
In contrast, SAM's semantic awareness allows it to generate superpixels that avoid these issues.

As shown in Tab.~\ref{tab:SLIC-SAM}, visual tokens derived from SAM-generated semantic superpixels significantly outperform those based on traditional SLIC superpixels.
To evaluate a more computationally efficient alternative, we then substituted the standard SAM with MobileSAM~\cite{MobileSAM}, a lightweight variant obtained through knowledge distillation, for both superpixel generation and decoding the [SEG] token.
While the reduced parameter count of MobileSAM results in a mild performance degradation, this approach still substantially surpasses both the SLIC-based method and compressive visual projectors in Tab.~\ref{tab:Compress-VS}.

\section{More Visualization Results}
\label{sec:more_viz}

\begin{figure}[t!]
    \centering
    \includegraphics[width=0.975\linewidth]{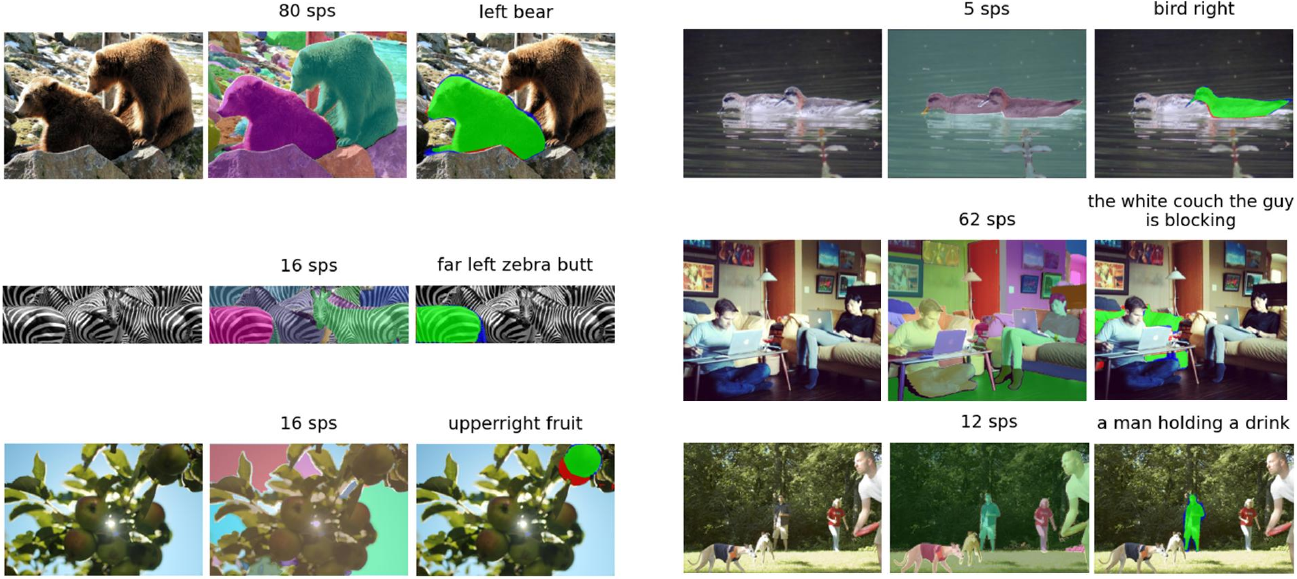}
    \captionof{figure}{
 RIS results obtained by our method. The arrangement aligns with Fig.~\ref{fig:main_viz}.
 }
    \label{fig:more_viz}
    \Description{}
\end{figure}

Fig.~\ref{fig:more_viz} presents additional RIS results generated by our method in challenging scenarios, including natural scenes, complex backgrounds, and small targets.

\end{document}